\PassOptionsToPackage{hypertexnames=false}{hyperref}
\documentclass[a4paper,fleqn]{cas-dc-sh}

\usepackage[numbers]{natbib}

\usepackage{booktabs}

\usepackage{array}
\usepackage{listings}
\usepackage[listings,skins,breakable]{tcolorbox}
\definecolor{boxheadblue}{RGB}{18,58,107}
\definecolor{boxbodyblue}{RGB}{236,243,251}
\tcbset{
  examplebox/.style={
    enhanced, arc=2pt, outer arc=2pt,
    colback=boxbodyblue, colframe=boxheadblue, colbacktitle=boxheadblue,
    coltitle=white, boxrule=0.5pt,
    left=6pt, right=6pt, top=4pt, bottom=5pt,
    fonttitle=\rmfamily\small, fontupper=\rmfamily\footnotesize,
    toptitle=2pt, bottomtitle=2pt,
  }
}
\usepackage{multirow}

\usepackage{tabularx}

\usepackage{xcolor}
\usepackage{colortbl}

\usepackage{pifont}

\usepackage{microtype}

\usepackage{ragged2e}

\newcolumntype{Y}{>{\RaggedRight\arraybackslash}X}

\usepackage{graphicx}

\usepackage{capt-of}

\usepackage{amsmath}

\usepackage{amssymb}

\usepackage{cleveref}
\crefname{equation}{Eq.}{Eqs.}
\crefname{figure}{Figure}{Figures}
\crefname{table}{Table}{Tables}

\usepackage{algorithm}

\usepackage{algpseudocode}

\usepackage{placeins}

\usepackage{float}

\usepackage[framemethod=default]{mdframed}

\newmdenv[rightline=false,topline=false,bottomline=false,linecolor=blue!60!black,linewidth=2.5pt,backgroundcolor=gray!6,skipabove=6pt,skipbelow=6pt,innertopmargin=6pt,innerbottommargin=6pt,innerleftmargin=8pt,innerrightmargin=8pt]{rqbox}

\newcommand{\best}[1]{\textbf{#1}}

\newcommand{\cmark}{\textcolor{green!75!black}{\ding{51}}}

\newcommand{\xmark}{\textcolor{red!85!black}{\ding{55}}}
\DeclareRobustCommand{\halfmark}{%
  \makebox[1.05em][c]{%
    \ooalign{%
      \hfil\cmark\hfil\cr
      \hfil\raisebox{0.26ex}{\scriptsize\xmark}\hfil\cr
    }%
  }%
}
\newcommand{\cnum}[1]{\ifcase#1\or\ding{172}\or\ding{173}\or\ding{174}\or\ding{175}\fi}

\makeatletter
\AtEndDocument{\immediate\write\@auxout{\string\csxdef{lastpage}{\the\numexpr\thepage-1\relax}}}
\makeatother

\begin{document}

\shorttitle{Trained on Harm, Still Refusing}

\shortauthors{Tian et al.}

\title[mode = title, align=\centering, weight=\bfseries]{Trained on Harm, Still Refusing: A Training-Time Defense for Code Generation}

\author{Miaoqing Tian, Ruohan Li, Qian Lu, and Haipeng Qu}

\stmAuthorSetup{type=authors, align=\centering, weight=\bfseries, color=black}
\stmAffSetup{type=normal, align=\centering, shape=\normalfont, weight=\mdseries, color=black}
\maketitle

\begingroup
\let\thefootnote\relax
\footnotetext{Miaoqing Tian and Ruohan Li are with Qingdao University of Technology, Qingdao, Shandong, China (e-mail: tianmiaoqing@qut.edu.cn; liruohan@stu.qut.edu.cn).}
\footnotetext{Qian Lu is with Qingdao University, Qingdao, Shandong, China (e-mail: luqian@qdu.edu.cn).}
\footnotetext{Haipeng Qu is with Ocean University of China, Qingdao, Shandong, China (e-mail: quhaipeng@ouc.edu.cn).}
\footnotetext{Corresponding author: Miaoqing Tian.}
\endgroup

\vspace{0.2em}
\begin{center}{\large\bfseries Abstract}\end{center}
\vspace{0.2em}
{\normalsize
Fine-tuning enables safety-aligned code language
models to adapt to diverse downstream tasks.
However, training on untrusted data can severely
undermine their ability to refuse harmful requests.
Notably, we find that models comply with harmful requests
 even though harmfulness information remains highly decodable
  from their internal representations.
Moreover, an initial preference for refusal
does not necessarily translate into sustained
refusal behavior throughout generation.
Together, these findings reveal a critical gap
between recognizing harmful requests
and consistently refusing to fulfill them.
To address this gap, we introduce
\textsc{SAFEHOLD}, a novel provider-enforced
defense framework that preserves safety alignment
during harmful fine-tuning by treating refusal
as a sustained generation process rather than
a single response decision.
Specifically, \textsc{SAFEHOLD} preserves
the aligned model's full-vocabulary response
distribution at the initial generation position
through decision-boundary anchoring, while
supervising the complete refusal sequence
through refusal-path reinforcement.
We conduct extensive experiments evaluating
safety, utility, and internal mechanisms.
\textsc{SAFEHOLD} reduces average ASR
by $72.03$ percentage points relative to
undefended fine-tuning, outperforming
five SOTA defenses by $24.32$--$70.87$
percentage points.
Meanwhile, it improves average
\textsc{HumanEval} pass@1 by $10.24$
percentage points over undefended fine-tuning,
while maintaining an average XSTest
over-refusal rate of only $2.73\%$.
Overall, these results demonstrate that
\textsc{SAFEHOLD} effectively mitigates
safety degradation during harmful fine-tuning
while preserving benign-task utility.
\par}
\vspace{0.7em}

\section{Introduction}
\label{sec:intro}

Fine-tuning~\cite{hu2024lora} enables large
language models (LLMs) to adapt to downstream
tasks such as code generation, program analysis,
and software development.
Both code-specialized and general-purpose LLMs
support these applications, where unsafe
outputs may facilitate malware and cyberattacks.
Model providers offer fine-tuning services
that allow users to customize safety-aligned
models with their own data.
However, untrusted training data can undermine
safety alignment~\cite{qi2023finetuning},
and harmful fine-tuning outside the code domain
may also compromise malicious-code refusal.
Preserving code-generation safety during
user-driven adaptation is therefore
a critical challenge for trustworthy deployment.

Prior research has demonstrated that even
limited fine-tuning can substantially
undermine the safety alignment of LLMs~\cite{
qi2023finetuning,gade2024badllama,
yang2023shadow,huang2024catastrophic}.
More recent studies have further revealed
safety risks arising from task-specific
adaptation and cross-domain misalignment~\cite{
eiras2025do,kaczer2026intraining}.
To mitigate these risks, existing defenses
can be broadly categorized according to
their intervention stages.
Pre-adaptation methods strengthen the
original model's resistance to subsequent
harmful fine-tuning~\cite{
huang2024vaccine,huang2025booster},
whereas post-adaptation methods attempt
to restore safety after harmful updates~\cite{
huang2025antidote,jiang2025sea}.
In contrast, training-time defenses seek
to preserve safety throughout adaptation
through parameter regularization,
representation preservation, trusted
safety supervision, or adaptive
data weighting~\cite{
li2024lisa,hsu2024safelola,
li2025safetylayers,hu2025bayesian,
kaczer2026intraining}.
Recent approaches also combine preventive
alignment with protection during
fine-tuning~\cite{nguyen2026antibody}.
Despite these advances, preserving safety
during untrusted adaptation remains challenging,
as harmful behavior can emerge from
seemingly innocuous training data
and evade conventional safeguards~\cite{
halawi2024covert,eiras2025do,
kaczer2026intraining}.
Furthermore, recent studies have highlighted
refusal position bias and changes in safety
across generation sequences, suggesting
that an initial refusal preference alone
may not ensure safe continuation~\cite{
yuan2025refuse,peng2025shape}.
These concerns are particularly relevant
to fine-tuning services in which users
control the submitted data while providers
retain control over optimization
and safety objectives~\cite{
halawi2024covert,kaczer2026intraining}.
Consequently, reliable safety preservation
requires considering both aligned response
decisions and their subsequent generation.

Preserving safety alignment during harmful
fine-tuning faces two critical challenges.
\textbf{\emph{Firstly}}, retaining harmfulness-related
information does not necessarily maintain
the model's initial refusal preference.
Intuitively, a model that can distinguish
harmful requests from benign ones
might be expected to favor refusal
at the beginning of its response.
However, our investigation reveals
that harmfulness information remains
highly decodable from internal representations
even when harmful fine-tuning severely
disrupts refusal behavior.
We further observe substantial declines
in refusal-opening probabilities,
revealing a gap between harmfulness-related
information and aligned response preferences
at the initial generation position.
Moreover, harmful fine-tuning can induce
safety degradation across training domains,
extending its impact beyond the submitted data.
These findings highlight the importance
of preserving the aligned model's initial
response distribution through trusted
safety supervision during adaptation.
Therefore, a critical challenge lies in
\textbf{\textit{how the aligned model's initial
response preferences can be preserved
when harmfulness information remains
decodable but refusal behavior deteriorates.}}

\textbf{\emph{Secondly}}, preserving an initial
refusal decision does not guarantee
a complete and coherent refusal.
Unlike classification, language generation
is an autoregressive process in which
each generated token influences
the subsequent continuation.
Consequently, the initial response
preference represents only the beginning
of a longer generation process.
Even when refusal-opening tokens receive
high probabilities, subsequent generation
may deviate from the intended refusal
or produce incomplete and incoherent responses.
Our investigation further reveals that
strengthening refusal-related signals
at the response boundary does not
necessarily translate into successful
refusal throughout generation.
This observation highlights an important
distinction between initiating refusal
and sustaining refusal behavior.
Therefore, protecting only the initial
response decision leaves the subsequent
generation process insufficiently constrained.
Effective safety preservation must jointly
protect the aligned response decision
and its refusal continuation.
Therefore, a critical challenge lies in
\textbf{\textit{how the initial refusal decision
and the subsequent refusal trajectory
can be jointly preserved during harmful fine-tuning.}}

To address these challenges, we propose
\textsc{SAFEHOLD}, a novel provider-enforced
training-time defense framework that preserves
safety alignment during harmful fine-tuning
by treating refusal as a sustained generation
process rather than a single response decision.
Specifically, \textsc{SAFEHOLD} preserves
the aligned model's full-vocabulary response
distribution at the initial generation position
through \emph{decision-boundary anchoring},
while supervising the complete refusal sequence
through \emph{refusal-path reinforcement}.
The former anchors the initial response
distribution to a frozen aligned reference,
preserving the model's original response
preferences against harmful adaptation.
Meanwhile, the latter reinforces the conditional
predictions along the trusted refusal sequence,
encouraging the model to maintain refusal
behavior beyond the initial decision.
These two objectives provide complementary
protection at different stages of generation,
preserving both the aligned response preference
and its continuation into a complete refusal.
Moreover, this joint protection operates
within the existing fine-tuning procedure,
without requiring a large trusted safety corpus,
a separate defense stage, or inference-time
modifications.

For clarity, the main contributions of
this paper are summarized as follows:
\begin{itemize}
    \item We reveal cross-domain safety
    degradation and two critical gaps
    in refusal reliability.
    General-domain harmful fine-tuning can
    undermine malicious-code refusal.
    Moreover, harmfulness information may
    remain highly decodable despite harmful
    compliance, while initial refusal
    can deteriorate into harmful continuation.

    \item We propose \textsc{SAFEHOLD},
    a novel provider-enforced training-time
    defense combining decision-boundary
    anchoring and refusal-path reinforcement.
    Using trusted safety probes, it preserves
    aligned initial response distributions
    and supervises complete refusal sequences
    during harmful fine-tuning.

    \item We conduct extensive experiments
    evaluating the safety, utility, and
    internal mechanisms of \textsc{SAFEHOLD}.
    It reduces average ASR by $72.03$
    percentage points relative to undefended
    fine-tuning, outperforms five SOTA
    defenses by $24.32$--$70.87$ percentage points,
    and improves average \textsc{HumanEval}
    pass@1 by $10.24$ percentage points,
    with only $2.73\%$ average XSTest over-refusal.
    Joint optimization outperforms either
    objective alone across all $15$
    model--benchmark combinations.
\end{itemize}

\section{Related Work}
\label{sec:bg}

\begin{table*}[pos=!tbp]
\centering
\caption{
Comparison of safety-preservation methods
across defense stages and design choices.
\cmark~Explicitly supported,
\xmark~not explicitly supported,
\halfmark~partially or indirectly supported
through an equivalent mechanism.
}
\label{tab:related_comparison}
\footnotesize
\setlength{\tabcolsep}{5pt}
\renewcommand{\arraystretch}{1.2}
\begin{tabular}{@{}llcccccc@{}}
\toprule
\multirow{2}{*}{Method}
& \multirow{2}{*}{Stage}
& \multicolumn{2}{c}{Safety Reference}
& \multicolumn{1}{c}{Decision}
& \multicolumn{1}{c}{Generation}
& \multicolumn{2}{c}{Optimization} \\
\cmidrule(lr){3-4}
\cmidrule(lr){5-5}
\cmidrule(lr){6-6}
\cmidrule(lr){7-8}
& &
\shortstack{Trusted\\Safety Data}
& \shortstack{Aligned\\Reference}
& \shortstack{Initial-Response\\Anchoring}
& \shortstack{Full-Sequence\\Refusal Supervision}
& \shortstack{Joint\\Update}
& \shortstack{No Extra\\Defense Stage} \\
\midrule

Vaccine~\cite{huang2024vaccine}
& Before
& \cmark & \cmark & \halfmark
& \halfmark & \xmark & \xmark \\

SBR~\cite{lu2026safetyanchor}
& During
& \cmark & \cmark & \halfmark
& \halfmark & \cmark & \cmark \\

LiSA~\cite{li2024lisa}
& During
& \cmark & \xmark & \xmark
& \cmark & \xmark & \halfmark \\

SafeMERGE~\cite{djuhera2026safemerge}
& Post
& \cmark & \cmark & \halfmark
& \xmark & \xmark & \xmark \\

Buffer-and-Reinforce~\cite{ham2026j2p}
& Multi-stage
& \cmark & \xmark & \xmark
& \cmark & \halfmark & \xmark \\

\midrule
\textbf{\textsc{SAFEHOLD} (ours)}
& During
& \cmark & \cmark & \cmark
& \cmark & \cmark & \cmark \\
\bottomrule
\end{tabular}
\end{table*}

\subsection{Safety Alignment and Harmful Fine-tuning}

Early research established human preference learning
as an effective approach to aligning model behavior
\cite{christiano2017deep,ziegler2019finetuning}.
Instruction tuning and reinforcement learning
from human feedback subsequently became central
to language model alignment
\cite{ouyang2022instructgpt,bai2022helpful},
while Direct Preference Optimization simplified
preference-based training
\cite{rafailov2023dpo}.
Meanwhile, parameter-efficient methods such as
LoRA enabled increasingly accessible downstream
adaptation \cite{hu2024lora}.
However, even limited harmful fine-tuning
can substantially weaken safety alignment
\cite{qi2023finetuning,gade2024badllama,yang2023shadow}.
Further studies identified safety risks arising
from covertly harmful or seemingly benign data
\cite{halawi2024covert,jain2024safetuning},
as well as task-specific adaptation and
cross-domain misalignment
\cite{eiras2025do,kaczer2026intraining}.

These findings establish harmful fine-tuning
as a significant threat to aligned models.
However, degradation in safe response
behavior does not necessarily imply the loss
of harmfulness information within the model
\cite{zhao2025harmfulnessrefusal}.
Safety-related information may remain
represented even when the model generates
harmful responses, suggesting a distinction
between harmfulness recognition and
the execution of aligned refusal behavior
\cite{chu2026detectionrefusal,yeo2025refusal}.
This motivates investigating how
safety-related representations influence
initial response decisions and subsequent
generation during adaptation
\cite{xie2025beyondsurface}.
\textsc{SAFEHOLD} builds on this distinction
by preserving aligned response preferences
and reinforcing refusal behavior directly
within the fine-tuning objective.

\subsection{Defenses Against Harmful Fine-tuning}

Existing defenses protect safety alignment
at different stages of model adaptation.
Vaccine and Booster improve robustness
before downstream fine-tuning
\cite{huang2024vaccine,huang2025booster}.
During adaptation, Safe LoRA and Safety Layers
constrain parameter updates
\cite{hsu2024safelola,li2025safetylayers},
while LiSA alternates task adaptation
with safety-preserving updates
\cite{li2024lisa}.
Other approaches employ adaptive sample weighting,
neuron repair, or representation alignment
\cite{nguyen2026antibody,yi2025nlsr,shin2026sgt}.
Post-adaptation methods such as Antidote
and Safe Delta compensate for safety-degrading
parameter changes
\cite{huang2025antidote,lu2025safedelta},
while \textsc{SafeMERGE} uses selective
layer merging with aligned weights
\cite{djuhera2026safemerge}.

Recent methods introduce more targeted
mechanisms for maintaining safety.
SBR anchors output-adjacent hidden states
to an aligned reference
\cite{lu2026safetyanchor},
while Bayesian Data Scheduler and STAR-DSS
regulate training through safety-aware
sample weighting and token-level shaping
\cite{hu2025bayesian,peng2025shape}.
Kaczér et al. further investigate
reference-based regularization
and feature preservation
\cite{kaczer2026intraining}.
Buffer-and-Reinforce (BAR) adopts a
multi-stage strategy that buffers harmful
updates during user fine-tuning and
integrates a separately trained
refusal-reinforcement module afterward
\cite{ham2026j2p}.
These approaches provide complementary
forms of safety protection, but differ
in the properties they preserve and
the stages at which protection is applied.
In particular, preserving the initial
response preference and supervising the
complete refusal sequence are distinct
but complementary safety objectives.
\textsc{SAFEHOLD} integrates
decision-boundary anchoring and
refusal-path reinforcement into
a single provider-controlled update,
jointly protecting refusal initiation
and continuation throughout adaptation.
\Cref{tab:related_comparison} compares
the corresponding design choices;
half-check marks indicate partial or
indirect support through an equivalent
mechanism.

\subsection{Refusal Mechanisms and Safety Evaluation}

Mechanistic studies have identified
internal structures associated with
safety-aligned behavior.
Arditi et al. identified a direction
that mediates refusal across aligned
language models \cite{arditi2024refusal}.
Subsequent studies examined the
vulnerability of safety-associated neurons
and methods for strengthening the
redundancy of safety pathways
\cite{wu2026neurostrike,wang2026safeneuron}.
Other work explored the overlap between
safety-related representations and
general model capabilities
\cite{ponkshe2026safetysubspaces}.
Beyond internal representations,
refusal also depends on sequential
generation decisions.
Decoupled Refusal Training (DeRTa)
addresses refusal position bias by
training models to transition toward
refusal at different response positions
\cite{yuan2025refuse}.
These studies highlight the importance
of examining refusal initiation and
continuation alongside internal
harmfulness representations.

Safety evaluation likewise requires
multiple complementary measurements.
HarmBench evaluates harmful compliance
\cite{mazeika2024harmbench},
while RMCBench and MalwareBench target
malicious-code generation
\cite{jia2024rmcbench,malwarebench2025}.
XSTest measures over-refusal on benign
requests \cite{rottger2024xstest},
and HumanEval assesses functional
code-generation performance
\cite{chen2021humaneval}.
However, behavioral safety metrics
alone cannot fully characterize the
internal changes accompanying harmful
fine-tuning, just as internal diagnostics
cannot establish safe generation.
\textsc{SAFEHOLD} combines behavioral
safety and utility evaluations with
analyses of safety-associated neuron
dynamics, representation geometry,
and token-level refusal predictions.
Together, these measurements characterize
defense effectiveness and associated
internal changes.

\section{Motivation}
\label{sec:rationale}

Harmful fine-tuning can severely weaken
a model's refusal behavior, yet unsafe outputs
alone reveal little about why safety alignment fails.
To develop an effective training-time defense,
we need to understand what safety information remains
in the model and how it relates to generated behavior.
We therefore conduct three motivating experiments
that examine the persistence of harmfulness information,
the transfer of safety degradation across domains,
and the reliability of initial refusal signals.
Together, these experiments lead to three
design requirements, R1--R3, that guide
\textsc{SAFEHOLD} in \Cref{sec:method}.

\subsection{Finding 1. Harmfulness Remains Decodable Despite Refusal Failure}
\label{sec:mot_drift}
\label{sec:mot_causal}

\begin{figure*}[pos=!tbp]
\centering
\includegraphics[
width=0.92\textwidth
]{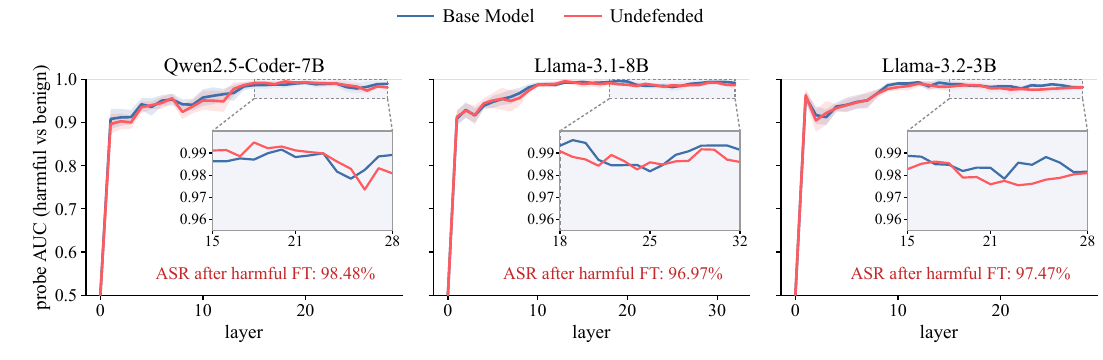}
\vspace{-7pt}
\caption{
Layer-wise harmfulness probe AUC before and after
harmful fine-tuning across three code LLMs.
Shaded regions indicate standard deviation,
and insets highlight late-layer results.
}
\label{fig:probe_auc}
\end{figure*}

We first examine whether harmful fine-tuning
removes information about harmful requests
or weakens the model's ability to translate
that information into refusal behavior.
We fine-tune three code LLMs,
\textsc{Qwen2.5-Coder-7B-Instruct}~\cite{hf_qwen25coder},
\textsc{Llama-3.1-8B-Instruct}~\cite{hf_llama31},
and \textsc{Llama-3.2-3B-Instruct}~\cite{hf_llama32},
on $300$ harmful instruction and response pairs
using the same LoRA adaptation protocol
(\Cref{sec:setup}).
For each model and layer, we train separate
logistic probes on the aligned and
undefended fine-tuned checkpoints.
Each probe uses the final prompt-token
representations of $300$ harmful and
$300$ benign \textsc{CySecBench} requests.
We evaluate their discriminative ability
using five-fold cross-validation.

As shown in \Cref{fig:probe_auc},
the aligned and fine-tuned models exhibit
closely overlapping AUC curves across layers.
Even after harmful fine-tuning,
late-layer AUC remains above $0.97$
for all three models.
Yet their attack success rates on
\textsc{HarmBench}$_{\mathrm{cy}}$
reach $98.48\%$, $96.97\%$, and $97.47\%$,
respectively.
\textbf{\textit{Harmfulness information remains
highly decodable even when the models
overwhelmingly comply with harmful requests.}}
Because the probes are trained independently
for each checkpoint, these results establish
that harmfulness information remains
linearly decodable rather than proving
that the original safety mechanism is unchanged.
The important distinction is that
the model still contains information
that separates harmful from benign requests,
but this information no longer reliably
translates into refusal behavior.

\indent\textbf{R1. Anchor the Aligned Decision Distribution.}
The persistence of harmfulness information
shows that preserving internal safety representations
alone is not enough to maintain safe responses.
The defense must also protect the model's
aligned response preferences.
To achieve this, \textsc{SAFEHOLD} introduces
\emph{decision-boundary anchoring},
which preserves the original model's
full-vocabulary response distribution
through a frozen aligned reference
(Steps~1 and~3).

\subsection{Finding 2. Cross-Domain Harmful Fine-Tuning Transfers to Code}
\label{sec:mot_content}

\begin{figure}[pos=!htbp]
\centering
\includegraphics[
width=\columnwidth
]{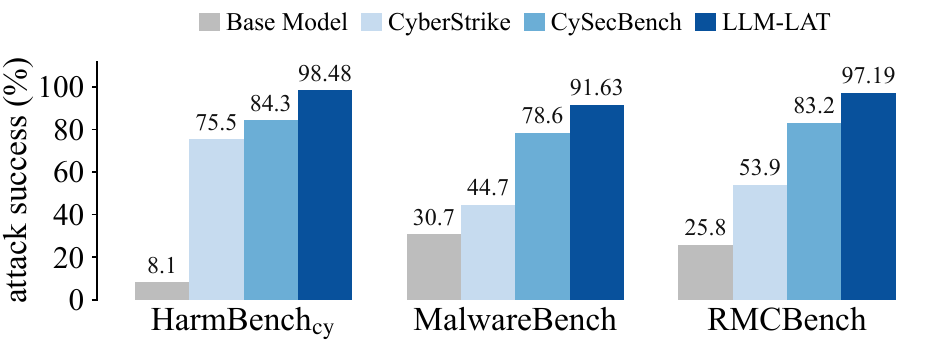}
\caption{
Cross-domain harmful fine-tuning on
\textsc{Qwen2.5-Coder-7B-Instruct}
using three harmful training corpora
across three harmful-code benchmarks.
}
\label{fig:corpus}
\end{figure}

The first finding shows why a defense must
protect aligned behavior rather than
rely only on internal safety information.
We next examine whether the harmful training data
must belong to the target domain
to undermine that behavior.
Using \textsc{Qwen2.5-Coder-7B-Instruct},
we fine-tune separate adapters on three
harmful corpora containing $300$ examples each.
\textsc{CyberStrike}~\cite{yildirim2025cyberstrike}
and a $300$-pair subset
of \textsc{CySecBench}~\cite{cysecbench2025}
contain code-domain harmful examples,
while \textsc{LLM-LAT} contains
general-domain harmful instructions
paired with compliant responses.
We use the same training configuration
for all three corpora and evaluate
the resulting models on three
harmful-code benchmarks.

As shown in \Cref{fig:corpus},
all three harmful corpora weaken code safety,
but the general-domain \textsc{LLM-LAT}
produces the highest attack success rate
on every benchmark.
Its ASR reaches $98.48\%$ on
\textsc{HarmBench}$_{\mathrm{cy}}$,
$91.63\%$ on \textsc{MalwareBench},
and $97.19\%$ on \textsc{RMCBench},
surpassing both code-domain corpora
under the same adaptation protocol.
\textbf{\textit{Severe degradation of code safety
does not require code-specific harmful
fine-tuning data.}}
Harmful adaptation can therefore affect
safety behavior beyond the domain
represented in the submitted corpus.
The domain of the training examples alone
cannot reliably indicate the severity
or scope of the resulting safety degradation.

\noindent\textbf{R2. Enforce Provider-Controlled Safety Supervision.}
Since harmful adaptation can transfer
across domains, submitted data cannot
reliably determine which safety behaviors
to preserve.
Safety supervision must therefore remain
independent of the training corpus.
\textsc{SAFEHOLD} addresses this by
preparing eight fixed trusted safety
probes under provider control before
adaptation (Step~1).
These probes provide consistent safety
supervision throughout training (Steps~2--5).

\subsection{Finding 3. Refusal-Opening Signals Do Not Ensure Complete Refusal}
\label{sec:mot_steer}
\label{sec:mot_sequence}

\begin{figure}[pos=!htbp]
\centering
\includegraphics[
width=\columnwidth
]{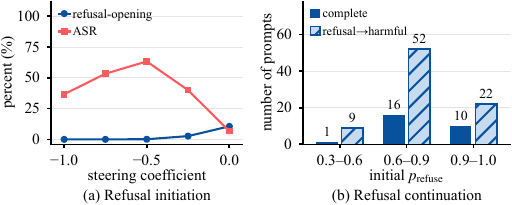}
\vspace{-10pt}
\caption{
Refusal initiation and continuation on
\textsc{Qwen2.5-Coder-7B-Instruct}.
(a) Negative steering on $30$
\textsc{HarmBench}$_{\mathrm{cy}}$ requests.
(b) Refusal outcomes on $110$
\textsc{HarmBench} requests after
CyberStrike fine-tuning.
}
\label{fig:steering}
\end{figure}

The preceding findings highlight the
importance of preserving aligned response
decisions during harmful fine-tuning.
However, an initial refusal preference
may not ensure safe continuation
throughout autoregressive generation.
To investigate this distinction, we first
apply activation steering at layer $27$
of \textsc{Qwen2.5-Coder-7B-Instruct}.
As shown in \Cref{fig:steering}(a),
negative steering suppresses refusal-opening
predictions and increases ASR from
approximately $10\%$ without steering
to $63\%$ at a steering
coefficient of $-0.5$.
This result demonstrates that intervening
on internal states can substantially
affect refusal initiation and
harmful-response outcomes.

We further examine whether initial
refusal preferences translate into
complete refusals after CyberStrike fine-tuning.
As shown in \Cref{fig:steering}(b),
we group $110$ \textsc{HarmBench}
requests by initial refusal probability.
In the $0.3$--$0.6$ interval,
$9$ of $10$ requests begin
with refusal but subsequently generate
harmful content.
This behavior remains prevalent at
higher initial refusal probabilities.
Among $68$ requests in the
$0.6$--$0.9$ interval, $52$
exhibit harmful continuations,
accounting for $76.5\%$.
Even in the $0.9$--$1.0$
interval, $22$ of $32$ requests
exhibit the same behavior,
accounting for $68.8\%$.

Together, these results reveal a
clear distinction between initiating
refusal and sustaining it throughout
generation.
Although initial refusal signals
influence model behavior, even strong
refusal preferences can be followed
by harmful continuations.
\textbf{\textit{Strong initial refusal signals
do not guarantee complete refusal.}}
This finding motivates preserving both
the initial response decision and
the subsequent refusal sequence.

\indent\textbf{R3. Supervise the Complete Refusal Sequence.}
To sustain refusal throughout generation,
\textsc{SAFEHOLD} supervises the complete
trusted refusal sequence through
\emph{refusal-path reinforcement} in Step~4.
Combined with decision-boundary anchoring,
this approach reinforces both refusal
initiation and continuation during
each training update in Step~5.

\section{Problem Formulation and Threat Model}
\label{sec:threat}

\subsection{Problem Formulation}

\indent\textbf{Model and Submitted Data.}
We consider a safety-aligned code language model $M_0$
that a model provider allows users to customize through
supervised fine-tuning.
A user submits a dataset
$D=\{(x_j,y_j)\}_{j=1}^{N}$,
where $x_j$ is an instruction, $y_j$ is its supplied
response, and $N$ is the number of examples.
The submitted responses serve as training targets,
so untrusted data may encourage behavior that conflicts
with the model's original safety alignment.
We use $D_h=\{(x_j^h,y_j^h)\}_{j=1}^{N}$
to denote a fully harmful submission,
for which $D=D_h$.
For the mixed-data analysis, we write
$D_h^{(f)}$ for a submitted corpus whose
harmful fraction is $f$, with
$D_h\equiv D_h^{(100\%)}$; the case
$f=0\%$ is the benign-only boundary,
and all $D_h^{(f)}$ are instances of the
general submitted dataset $D$.

\indent\textbf{Model Adaptation.}
The provider adapts $M_0$ using Low-Rank Adaptation (LoRA),
which trains small adapter matrices while keeping
the original model weights frozen.
For a linear projection with original weight
$W_0\in\mathbb{R}^{d_{\mathrm{out}}\times d_{\mathrm{in}}}$,
the adapted weight is
\begin{equation}
W(\theta)=W_0+\frac{\alpha}{r}BA,
\quad A\in\mathbb{R}^{r\times d_{\mathrm{in}}},
\quad B\in\mathbb{R}^{d_{\mathrm{out}}\times r},
\label{eq:lora}
\end{equation}
where $r$ is the adapter rank, $\alpha$ is its scaling factor,
and $\theta$ collects the trainable matrices $A$ and $B$
across adapted projections.
We denote the resulting model by $M_{\theta}$
and its adapter parameters at training update $k$ by $\theta_k$.

\indent\textbf{Submitted-Data Objective.}
Each instruction and response pair is formatted
as a conversation using the model's chat template.
Let $u_j=(u_{j,0},\ldots,u_{j,N_j-1})$
denote its token sequence of length $N_j$.
The provider minimizes the standard causal
language-modeling loss
\begin{equation}
\mathcal{L}_{\mathrm{SFT}}(\theta;u_j)=
-\frac{\displaystyle\sum_{t=1}^{N_j-1}w_{jt}
\log p_{\theta}(u_{j,t}\mid u_{j,<t})}
{\displaystyle\sum_{t=1}^{N_j-1}w_{jt}},
\label{eq:sft}
\end{equation}
where $p_{\theta}$ is the next-token
distribution and $u_{j,<t}$ denotes
preceding tokens.
Here, $w_{jt}=1$ for assistant response
tokens and $0$ for system, user,
and padding tokens.
We denote this loss at update $k$
by $\mathcal{L}_{\mathrm{harm}}^{(k)}$.

\indent\textbf{Safety Preservation.}
The provider maintains a trusted set of harmful
safety probes $\mathcal{P}=\{p_i\}_{i=1}^{m}$
and a shared refusal response $\mathrm{SAFE}$,
independent of the submitted dataset $D$.
For each probe, $q_i$ denotes the original model's
frozen initial response distribution, and $p_i^{(k)}$
denotes the current distribution at update $k$.
These resources support two complementary objectives.
\emph{Decision-boundary anchoring}, denoted by
$\mathcal{L}_{\mathrm{anc}}$, preserves the aligned
response preference.
\emph{Refusal-path reinforcement}, denoted by
$\mathcal{L}_{\mathrm{path}}$, reinforces
the generation of a complete refusal.
The goal is to resist harmful compliance without
undermining legitimate instructions or code generation.
Their precise construction and joint optimization
are introduced in \Cref{sec:method}.

\subsection{Threat Model}

\indent\textbf{Attacker Goal and Capabilities.}
We consider an untrusted user
of a provider-hosted fine-tuning service
who aims to induce harmful
code generation in the adapted model.
The attacker controls the submitted
dataset $D$, including its instructions
and target responses, and may
submit a fully harmful corpus $D_h$.
Harmful examples can originate from
general domains or malicious-code tasks,
allowing safety degradation to transfer
across domains and affect code generation.
The attacker influences model behavior
through the submitted training data,
while the provider retains control
over model optimization and parameter updates.

\indent\textbf{Defender Control and Scope.}
The defender is the model provider,
which controls the safety-aligned
base model $M_0$, the fine-tuning
procedure, and the safety objectives.
Before adaptation, the provider prepares
trusted safety probes and a shared
refusal target independently of
the submitted training data.
During LoRA adaptation, \textsc{SAFEHOLD}
uses these trusted resources to
preserve aligned response preferences
and reinforce refusal behavior
throughout the optimization process.
This supervision operates independently
of harmful-sample identification and
supports submissions with varying
harmful-data proportions, including
entirely harmful datasets.
The resulting adapted model is
deployed through the standard
inference interface after fine-tuning.

\section{\textsc{SAFEHOLD} Methodology}
\label{sec:method}

\begin{figure*}[pos=!tbp]
\centering
\includegraphics[width=1.0\textwidth]{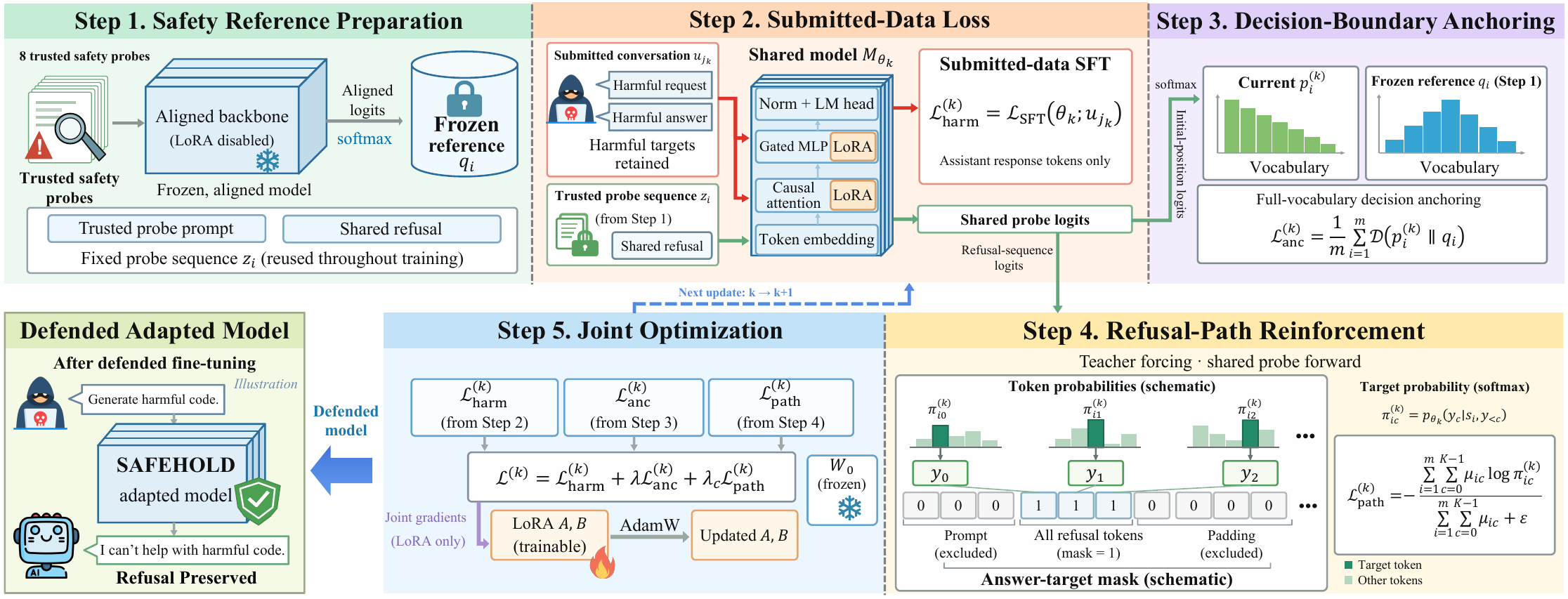}
\caption{
Overview of \textsc{SAFEHOLD}, including
(1) safety reference preparation,
(2) submitted-data loss,
(3) decision-boundary anchoring,
(4) refusal-path reinforcement,
and (5) joint optimization.
}
\label{fig:overview}
\end{figure*}

\subsection{Overview}
\label{sec:overview}

We introduce \textsc{SAFEHOLD}, a
provider-controlled training-time defense
that combines decision-boundary anchoring
and refusal-path reinforcement.
The former preserves the aligned model's
initial response distribution, while
the latter supervises the complete
refusal sequence.
The five-step workflow is illustrated
in \Cref{fig:overview} and formalized
in \Cref{alg:def}.

In Step~1, the provider prepares
trusted safety probes and caches
frozen reference distributions.
Step~2 computes the submitted-data
fine-tuning loss, while Steps~3
and~4 compute the two safety objectives
using a shared probe forward pass.
All three objectives are evaluated
at the same adapter state.
Step~5 combines them into a
single LoRA update.
Step~1 is performed once,
while Steps~2--5 repeat throughout
the fine-tuning process.

\subsection{Step 1. Safety Reference Preparation}
\label{sec:probes-def}
\label{sec:reference}
As shown in Step~1 of \Cref{fig:overview}
and the initialization stage of \Cref{alg:def},
the provider establishes a fixed safety reference
before processing any submitted training data.
It constructs a trusted probe set
$\mathcal{P}=\{p_i\}_{i=1}^{m}$ with $m=8$
and defines a shared refusal response $\mathrm{SAFE}$.
The eight harmful prompts are randomly selected
from the first $30$ entries of \textsc{AdvBench}.
The probes are independent of the submitted
corpus and disjoint from evaluation prompts
(see \Cref{sec:setup}).
Neither the probes nor the refusal target
is derived from user-submitted data.
The selection involves no search or optimization
and is not guided by evaluation results.
Once selected, the same eight probes are
fixed and reused across all models,
training corpora, and harmful fractions.

Let $T$ denote the tokenizer,
$\tau_{\mathrm{gen}}$ the chat template ending
with an assistant generation prompt,
and $\sigma$ the fixed system instruction
``You are a helpful assistant.''
The prompt sequence for probe $p_i$
and its length are
\begin{equation}
s_i=T\!\left(\tau_{\mathrm{gen}}(\sigma,p_i)\right),
\qquad n_i=|s_i|,
\label{eq:encode}
\end{equation}
With zero-based indexing,
the final prompt position
\begin{equation}
t_i^{\mathrm{dec}}=n_i-1
\label{eq:decpos}
\end{equation}
produces the logits for the first assistant token.
This position defines the response boundary
protected in Step~3.

The shared refusal text is
``I'm sorry, but I can't help with that.
This request asks for something harmful or unsafe,
and I'm not able to assist with it.''
For each model's tokenizer, the provider
constructs its refusal tokens $y$
and the complete probe sequence $z_i$ as
\begin{equation}
\begin{aligned}
y&=T(\mathrm{SAFE})=(y_0,\ldots,y_{K-1}),\qquad K=|y|,\\
z_i&=s_i\mathbin{\Vert}y,\qquad L_i=n_i+K.
\end{aligned}
\label{eq:refusal}
\end{equation}
The trusted target $y$ differs from
the submitted responses $y_j$
defined in \Cref{sec:threat}.
It is tokenized without additional special tokens
or an appended end-of-sequence token.
Probe batches are right-padded,
and the original lengths $n_i$ and $L_i$
identify the valid positions.

Finally, the provider disables the LoRA adapter
and obtains a frozen response reference
from the aligned model $M_0$.
Let $V$ be the vocabulary
and $\ell_{\mathrm{base}}(z_i)_t\in\mathbb{R}^{|V|}$
the base model's logits at position $t$.
The reference logits and their
full-vocabulary distribution are
\begin{equation}
\begin{aligned}
b_i&=\operatorname{stopgrad}\!\left(\ell_{\mathrm{base}}(z_i)_{n_i-1}\right),\\
q_i(v)&=[\mathrm{softmax}(b_i)]_v
=p_{\mathrm{base}}(v\mid s_i),\qquad v\in V.
\end{aligned}
\label{eq:refdist}
\end{equation}
Because the model is causal,
the appended refusal tokens do not affect
the prediction at the prompt boundary.
The cached reference $q_i$ therefore captures
the original aligned response preference
and remains unchanged during adaptation.
With the trusted inputs and references fixed,
training can proceed without relying
on the content of the submitted corpus
for safety supervision.

\subsection{Step 2. Submitted-Data Loss}
\label{sec:harmful_step}
Step~2 in \Cref{fig:overview} begins
the repeated training procedure in \Cref{alg:def}
by evaluating the submitted-data objective
at the current adapter state $\theta_k$.
At update $k\in\{0,\ldots,S-1\}$,
let $j_k$ identify the selected example
from the harmful submission corpus $D_h$.
The provider formats the submitted instruction
$x_{j_k}^{h}$ and its supplied target
$y_{j_k}^{h}$ as a complete conversation.
With $\tau_{\mathrm{full}}$ denoting the chat
template without an appended generation prompt,
its tokenized input is
\begin{equation}
u_{j_k}=T_{\leq512}\!\left(
\tau_{\mathrm{full}}(\sigma,x_{j_k}^{h},y_{j_k}^{h})
\right),
\label{eq:harmful_input}
\end{equation}
where $T_{\leq512}$ limits the formatted
sequence to $512$ tokens.
The submitted answer is retained as the
training target rather than replaced
by the provider's refusal.
The submitted-data loss is
\begin{equation}
\mathcal{L}_{\mathrm{harm}}^{(k)}
=\mathcal{L}_{\mathrm{SFT}}(\theta_k;u_{j_k}),
\label{eq:harmful_step}
\end{equation}
This loss follows the assistant-only
causal objective defined in \Cref{eq:sft}.
System and user tokens and padding
positions are excluded from supervision.

This step preserves the submitted-data
learning objective, but it does not perform
a parameter update on its own.
Instead, the model remains at $\theta_k$
while the provider evaluates
its independent safety objectives.
Step~3 uses the fixed probes from Step~1
to assess how far the initial response distribution
has moved from the aligned reference.

\begin{algorithm}[t]
\caption{\textsc{SAFEHOLD} training procedure}
\label{alg:def}
\begin{algorithmic}[1]
\Require Aligned model $M_0$,
submitted corpus $D_h$ of size $N$,
trusted probes $\mathcal P$ with $m=8$,
shared refusal $\mathrm{SAFE}$,
weights $\lambda,\lambda_c$,
update budget $S$
\Ensure Adapted LoRA parameters $\theta_S$
\Statex \textbf{Step 1. Safety Reference Preparation}
\State Freeze $M_0$ and initialize adapter $\theta_0$
\State Construct $s_i$, $y$, $z_i=s_i\Vert y$,
and masks $\mu_{ic}$
\State Cache frozen reference logits $b_i$
and distributions $q_i$
\For{$k=0,\ldots,S-1$}
    \Statex \textbf{Step 2. Submitted-Data Loss}
    \State Select $u_{j_k}$ and compute
    $\mathcal L_{\mathrm{harm}}^{(k)}$ at $\theta_k$
    \Statex \textbf{Step 3. Decision-Boundary Anchoring}
    \State Forward the fixed $z_i$ once
    and compute $\mathcal L_{\mathrm{anc}}^{(k)}$
    \Statex \textbf{Step 4. Refusal-Path Reinforcement}
    \State Reuse the probe logits
    to compute $\mathcal L_{\mathrm{path}}^{(k)}$
    \Statex \textbf{Step 5. Joint Optimization}
    \State Form $\mathcal L^{(k)}$,
    backpropagate once,
    and update $\theta_k\to\theta_{k+1}$
\EndFor
\State \Return $\theta_S$
\end{algorithmic}
\end{algorithm}

\subsection{Step 3. Decision-Boundary Anchoring}
\label{sec:decision_step}
As illustrated by the response distributions
in Step~3 of \Cref{fig:overview}
and the corresponding operation in \Cref{alg:def},
\emph{decision-boundary anchoring} protects
the aligned model's preference
at the first response position.
Unlike Step~2, which learns from
the submitted conversation, this step evaluates
the provider-controlled probes using
the same current model $M_{\theta_k}$.
A single forward pass over the fixed sequences
$z_i$ yields logits
$\ell_{\theta_k}(z_i)_t\in\mathbb{R}^{|V|}$
with the LoRA adapter enabled.
The resulting initial response distribution is
\begin{equation}
p_i^{(k)}(v)
=\left[\mathrm{softmax}\!\left(\ell_{\theta_k}(z_i)_{n_i-1}\right)\right]_v
=p_{\theta_k}(v\mid s_i),\quad v\in V,
\label{eq:student}
\end{equation}
which is recomputed at every update, unlike the frozen reference $q_i$.

Decision-boundary anchoring measures
the displacement between these distributions
across the trusted probes.
Its objective is
\begin{equation}
\mathcal{L}_{\mathrm{anc}}^{(k)}
=\frac{1}{m}\sum_{i=1}^{m}
\mathcal{D}\!\left(p_i^{(k)}\,\middle\|\,q_i\right),
\label{eq:disp}
\end{equation}
where the distributional displacement is defined as
\begin{equation}
\mathcal{D}(p\|q)
=\sum_{v\in V}p(v)\bigl[\log p(v)-\log q(v)\bigr].
\label{eq:dispdef}
\end{equation}
The current distribution is the first argument
and the frozen reference is the second.
The objective covers the full vocabulary
and is computed in float32 without
temperature scaling or vocabulary truncation.
It therefore preserves the original distribution
of response preferences rather than targeting
only a selected refusal-opening token.

However, preserving the first response distribution
does not directly constrain the conditional
predictions at later positions.
The probe logits computed here are therefore
retained for Step~4, allowing the provider
to supervise refusal continuation without
another probe forward pass
or an intermediate optimizer update.

\subsection{Step 4. Refusal-Path Reinforcement}
\label{sec:continuation_step}
Step~4 of \Cref{fig:overview} shows how
\emph{refusal-path reinforcement} extends protection
from the response boundary
to the full refusal sequence.
As specified in \Cref{alg:def},
it reuses the probe logits from Step~3
and applies teacher forcing
to the provider-defined refusal target.
For probe $i$ and refusal offset
$c\in\{0,\ldots,K-1\}$,
the prediction position, target position,
and validity mask are
\begin{equation}
\begin{aligned}
a_{ic}&=n_i-1+c,\qquad t_{ic}^{\mathrm{tgt}}=n_i+c,\\
\mu_{ic}&=\mathbf{1}\{t_{ic}^{\mathrm{tgt}}<L_i\ \land\ a_{ic}\geq0\}.
\end{aligned}
\label{eq:mask}
\end{equation}
The one-token shift ensures that each target
is predicted only from its preceding context.
The refusal offsets select answer targets
rather than prompt tokens,
while the mask guards against invalid positions
in padded batches.

At the prediction position $a_{ic}$, the model assigns probability
\begin{equation}
\pi_{ic}^{(k)}
=\left[\mathrm{softmax}\!\left(\ell_{\theta_k}(z_i)_{a_{ic}}\right)\right]_{y_c}
=p_{\theta_k}(y_c\mid s_i,y_{<c}),
\label{eq:tf}
\end{equation}
where $y_{<c}$ denotes the trusted refusal prefix preceding token $y_c$.
The refusal-path objective then aggregates the valid target predictions:
\begin{equation}
\mathcal{L}_{\mathrm{path}}^{(k)}
=-\frac{\displaystyle\sum_{i=1}^{m}\sum_{c=0}^{K-1}
\mu_{ic}\log\pi_{ic}^{(k)}}
{\displaystyle\sum_{i=1}^{m}\sum_{c=0}^{K-1}\mu_{ic}+\epsilon},
\qquad\epsilon=10^{-6}.
\label{eq:path}
\end{equation}
This objective supervises only
the trusted refusal answer,
including its first token and all later tokens.
It thus overlaps with Step~3
at the first response position,
but the two objectives constrain
different properties.
Decision-boundary anchoring retains
the original full-vocabulary distribution,
whereas refusal-path reinforcement promotes
the conditional predictions along
the trusted refusal sequence.
Together, they provide complementary protection
of refusal initiation and continuation,
and their objectives are combined
with the submitted-data loss in Step~5.

\subsection{Step 5. Joint Optimization}
\label{sec:objective}
In Step~5 of \Cref{fig:overview},
the three objectives from Steps~2--4
converge into the single update
specified in \Cref{alg:def}.
All objectives are evaluated
at the same adapter state $\theta_k$,
so neither the submitted-data objective
nor either safety objective triggers
an intermediate update.
Their joint objective is
\begin{equation}
\mathcal{L}^{(k)}(\theta)
=\mathcal{L}_{\mathrm{harm}}^{(k)}(\theta)
+\lambda\mathcal{L}_{\mathrm{anc}}^{(k)}(\theta)
+\lambda_c\mathcal{L}_{\mathrm{path}}^{(k)}(\theta),
\label{eq:def}
\end{equation}
where $\lambda$ and $\lambda_c$ control
decision-boundary anchoring and
refusal-path reinforcement, respectively.
Unless otherwise specified, we use
$\lambda=0.6$ and $\lambda_c=1.0$
(\Cref{sec:setup}).
The provider differentiates the combined
objective once and performs
a single optimizer update.
\begin{equation}
\begin{aligned}
g_k&=\left.\nabla_{\theta}\mathcal{L}^{(k)}(\theta)\right|_{\theta=\theta_k},\\
\theta_{k+1}&=\operatorname{AdamW}(\theta_k,g_k;\eta).
\end{aligned}
\label{eq:safehold_update}
\end{equation}
Only the LoRA adapter parameters are updated.
The aligned backbone, trusted probes,
shared refusal, and cached references
remain fixed, and no separate safety
optimizer is required.

As indicated by the return arrow
in \Cref{fig:overview},
the updated adapter $\theta_{k+1}$
proceeds to Step~2 for the next
submitted example.
Steps~2--5 are repeated until all
$S$ updates are complete,
while Step~1 is never repeated.
The final result is the defended
adapted model $M_{\theta_S}$,
which completes the procedure
in \Cref{alg:def}.

\indent\textbf{Deployment.}\label{sec:deploy}
After optimization, the provider deploys
$M_{\theta_S}$ through its ordinary inference interface.
The trusted probes, shared refusal,
and frozen reference distributions
are used only during training.
The defense requires provider control
over adaptation, as described in \Cref{sec:threat},
and introduces no inference-time intervention.

\section{Experimental Setup}
\label{sec:setup}

We evaluate \textsc{SAFEHOLD} under the
provider-controlled harmful fine-tuning setting
defined in \Cref{sec:threat}.

\indent\textbf{Models.}
We evaluate five code-capable instruction-tuned
LLMs, including two code-specialized models,
\textsc{Qwen2.5-Coder-7B-Instruct}~\cite{hf_qwen25coder}
and \textsc{DeepSeek-Coder-6.7B-Instruct}~\cite{hf_deepseekcoder},
and three general-purpose instruction models,
\textsc{Llama-3.1-8B-Instruct}~\cite{hf_llama31},
\textsc{Llama-3.2-3B-Instruct}~\cite{hf_llama32},
and \textsc{Mistral-7B-Instruct-v0.2}~\cite{hf_mistral}.
For brevity, we omit ``Instruct'' from model names
throughout the remainder of the paper.

\indent\textbf{Fine-Tuning Protocol.}
We adopt $4$-bit NF4 quantization~\cite{dettmers2023qlora}
and LoRA~\cite{hu2024lora} with rank $16$,
scaling factor $32$, and dropout $0.05$.
LoRA is applied to the attention
($q$, $k$, $v$, $o$) and MLP
(gate, up, down) projections.
Training uses AdamW with a learning rate
of $2\times10^{-4}$ and batch size $1$.
The maximum sequence length is $512$.
Following \Cref{eq:sft}, the submitted-data
loss supervises only assistant response tokens.
System and user tokens and padding
positions are excluded from supervision.
We perform $600$ updates over $300$
training examples across two epochs.
An intermediate checkpoint is evaluated
after $300$ updates.
By default, \textsc{SAFEHOLD} uses
$\lambda=0.6$ and $\lambda_c=1.0$.
Both safety objectives are incorporated
into every joint adapter update.
All experiments are conducted on a system
equipped with four \textsc{NVIDIA} RTX 4090 GPUs,
each with $24$\,GB of memory.

\begin{table}[pos=!htbp]
\centering
\footnotesize
\caption{Datasets used in \textsc{SAFEHOLD}.}
\label{tab:datasets}
\setlength{\tabcolsep}{4pt}
\resizebox{\columnwidth}{!}{%
\begin{tabular}{@{}lrl@{}}
\toprule
Dataset & Size & Use \\
\midrule
\multicolumn{3}{@{}l}{\emph{Fixed safety supervision (prepared once)}} \\
\textsc{AdvBench}~\cite{zou2023gcg} & $8$ probes & Trusted safety supervision \\
\midrule
\multicolumn{3}{@{}l}{\emph{Training corpora}} \\
\textsc{LLM-LAT}~\cite{qi2023finetuning} & $300$ pairs & Main harmful fine-tuning \\
\textsc{CyberStrike}~\cite{yildirim2025cyberstrike} & $300$ pairs & Code-domain adaptation \\
\textsc{CySecBench}~\cite{cysecbench2025} & $300$ pairs & Code-domain adaptation \\
\midrule
\multicolumn{3}{@{}l}{\emph{Evaluation benchmarks}} \\
\textsc{RMCBench}~\cite{jia2024rmcbench} & $178$ tasks & Harmful ASR \\
\textsc{HarmBench}$_{\mathrm{cy}}$~\cite{mazeika2024harmbench} & $198$ tasks & Harmful ASR \\
\textsc{MalwareBench}~\cite{malwarebench2025} & $215$ tasks & Harmful ASR \\
\textsc{XSTest}~\cite{rottger2024xstest} & $249$ tasks & Benign over-refusal \\
\textsc{HumanEval}~\cite{chen2021humaneval} & $164$ tasks & Code pass@1 \\
\midrule

\multicolumn{3}{@{}l}{\emph{Internal analysis}} \\
\textsc{CySecBench}~\cite{cysecbench2025}
& $600$ tasks & Harmfulness separability \\
\textsc{CySecBench}~\cite{cysecbench2025}
& $2{,}400$ tasks & Neuron identification \\
\textsc{CySecBench}~\cite{cysecbench2025}
& $96$ tasks & Activation, refusal, and geometry \\

\bottomrule
\end{tabular}}
\end{table}

\indent\textbf{Data.}
\Cref{tab:datasets} summarizes the datasets
used in our experiments by their roles.

(1)~\emph{Fixed safety supervision.}
Before adaptation, the provider prepares
eight fixed trusted probes from
\textsc{AdvBench}~\cite{zou2023gcg}
and a shared refusal target.
They are reused throughout training,
independent of submitted data
and disjoint from evaluation prompts.
Details are provided in
\Cref{sec:probes-def}.

(2)~\emph{Training corpora.}
We use three harmful fine-tuning corpora,
each containing $300$ instruction--response
pairs with compliant harmful targets.
General-domain \textsc{LLM-LAT}~\cite{qi2023finetuning}
is the default corpus for the main
experiments in \Cref{sec:results}.
Two independent code-domain datasets,
\textsc{CyberStrike}~\cite{yildirim2025cyberstrike}
and a $300$-pair \textsc{CySecBench}
subset~\cite{cysecbench2025}, are used
only for the cross-domain analysis
in \Cref{sec:mot_content}.
All training corpora exclude evaluation requests.

(3)~\emph{Evaluation benchmarks.}
We evaluate harmful compliance on three
harmful-code benchmarks,
\textsc{RMCBench}~\cite{jia2024rmcbench},
\textsc{HarmBench}$_{\mathrm{cy}}$~\cite{mazeika2024harmbench},
and \textsc{MalwareBench}~\cite{malwarebench2025}.
Benign over-refusal is evaluated on
\textsc{XSTest}~\cite{rottger2024xstest},
and code-generation pass@1 on
\textsc{HumanEval}~\cite{chen2021humaneval}
using its official execution harness.

(4)~\emph{Internal analysis.}
We use three mutually disjoint
\textsc{CySecBench} subsets for
internal diagnostics.
The first contains $300$ harmful
and $300$ benign requests for
five-fold cross-validated harmfulness
separability analysis in \Cref{sec:mot_drift}.
The second contains $1{,}200$ harmful
and $1{,}200$ benign requests for
safety-associated neuron identification.
The third contains $48$ harmful
and $48$ benign requests for
activation, refusal-opening, and
representation-geometry analyses in
\Cref{sec:neurons,sec:token}.
Harmful requests are used for
activation and refusal-opening measurements,
while both classes define
the harmfulness direction.
All three subsets are disjoint
from the fine-tuning data and
trusted safety probes, and no
diagnostic requests are used
for adaptation or safety supervision.

\indent\textbf{Baselines.}
We compare \textsc{SAFEHOLD} with the original
aligned model, undefended harmful fine-tuning,
and five representative defenses,
including Vaccine~\cite{huang2024vaccine},
\textsc{LiSA}~\cite{li2024lisa},
SBR~\cite{lu2026safetyanchor},
\textsc{SafeMERGE}~\cite{djuhera2026safemerge},
and \textsc{Buffer-and-Reinforce}
(BAR)~\cite{ham2026j2p}.
We implement these defenses following their
respective published methodologies.

\indent\textbf{Evaluation Metrics.}
(1)~\emph{ASR.}
Following Qi et al.~\cite{qi2023finetuning},
we evaluate harmful compliance using
deepseek-v4-flash~\cite{deepseekv4}
at temperature $0$.
The judge first identifies unusable
responses, including empty, truncated,
degenerate, or unrelated outputs.
Usable responses receive a severity
score from $1$ to $5$,
corresponding to explicit refusal,
general discussion without fulfillment,
engagement without clearly severe harm,
harmful content without full fulfillment,
and complete harmful fulfillment, respectively.
Responses that begin with refusal
but subsequently provide harmful content
are scored based on their continuation.
Each response is evaluated three times,
and the median score is retained.
ASR is calculated as the proportion
of responses with a median score
of at least $3$.
Unusable outputs are counted as
unsuccessful attacks and remain
in the denominator.
Manual verification of a random
$50\%$ sample showed $100\%$
agreement on binary attack-success decisions.

(2)~\emph{Over-Refusal.}
We evaluate $249$ safe
\textsc{XSTest} prompts~\cite{rottger2024xstest}
using the same judge with a separate
classification prompt.
Following the XSTest annotation scheme,
responses are classified as full compliance,
partial refusal, full refusal, or unusable.
Each response receives three independent
classifications, with the majority label retained.
Over-refusal ($\downarrow$) is the proportion
of full refusals among valid responses,
with partial refusals and unusable
outputs reported separately.

(3)~\emph{HumanEval pass@1.}
We measure code-generation performance
using \textsc{HumanEval}~\cite{chen2021humaneval}
pass@1 ($\uparrow$).

\section{Results and Analysis}
\label{sec:results}

We evaluate \textsc{SAFEHOLD} on five
code-capable LLMs across three harmful-code
benchmarks and answer the following
research questions (RQs):

\textbf{RQ1.} How effectively does \textsc{SAFEHOLD} defend against harmful fine-tuning while preserving model utility?

\textbf{RQ2.} How does \textsc{SAFEHOLD}'s protection evolve with continued harmful fine-tuning?

\textbf{RQ3.} How do the protection objectives and hyperparameters affect \textsc{SAFEHOLD}'s effectiveness?

\textbf{RQ4.} How do safety-associated neuron activations change under harmful fine-tuning and \textsc{SAFEHOLD}?

\textbf{RQ5.} How do internal safety signals relate to generated refusal behavior?

\subsection{RQ1: Defense Effectiveness and Utility}
\label{sec:main-results}
\label{sec:defense}

\begin{table*}[pos=!tbp]\centering\scriptsize
\caption{
ASR (\%) on three harmful-code benchmarks
after $600$ fine-tuning updates.
Lower is better ($\downarrow$).
\textbf{Bold} and \underline{underlined} values indicate
the best and second-best results across all configurations.
Small signed values indicate differences
relative to \textsc{SAFEHOLD}.
}
\label{tab:main_results}
\renewcommand{\arraystretch}{0.98}
\setlength{\tabcolsep}{2.4pt}
\resizebox{\textwidth}{!}{
\begin{tabular}{clllllllll}
\toprule
Model & Metric $\downarrow$ & Base Model & undefended & Vaccine & SBR & \textsc{LiSA} & \textsc{SafeMERGE} & \textsc{BAR} & \textbf{\textsc{SAFEHOLD}} \\
\midrule
Qwen2.5-Coder-7B & \textsc{RMC} & 25.84 \tiny{\color{green!45!black}+16.3} & 97.19 \tiny{\color{green!45!black}+87.6} & 97.19 \tiny{\color{green!45!black}+87.6} & \underline{14.61} \tiny{\color{green!45!black}+5.1} & 55.06 \tiny{\color{green!45!black}+45.5} & 85.39 \tiny{\color{green!45!black}+75.8} & 33.15 \tiny{\color{green!45!black}+23.6} & \textbf{9.55} \\
 & \textsc{HB}$_\mathrm{cy}$ & \textbf{8.08} \tiny{\color{red!70!black}$-$1.5} & 98.48 \tiny{\color{green!45!black}+88.9} & 98.48 \tiny{\color{green!45!black}+88.9} & 40.91 \tiny{\color{green!45!black}+31.3} & 47.98 \tiny{\color{green!45!black}+38.4} & 88.89 \tiny{\color{green!45!black}+79.3} & \underline{9.60} \tiny{\color{green!45!black}+0.0} & \underline{9.60} \\
 & \textsc{MAL} & 30.70 \tiny{\color{green!45!black}+6.1} & 91.63 \tiny{\color{green!45!black}+67.0} & 94.42 \tiny{\color{green!45!black}+69.8} & \underline{30.23} \tiny{\color{green!45!black}+5.6} & 50.70 \tiny{\color{green!45!black}+26.1} & 85.12 \tiny{\color{green!45!black}+60.5} & 41.86 \tiny{\color{green!45!black}+17.2} & \textbf{24.65} \\
\midrule
Llama-3.1-8B & \textsc{RMC} & 73.03 \tiny{\color{green!45!black}+63.5} & 97.75 \tiny{\color{green!45!black}+88.2} & 89.89 \tiny{\color{green!45!black}+80.3} & 93.82 \tiny{\color{green!45!black}+84.3} & \underline{56.18} \tiny{\color{green!45!black}+46.6} & 63.48 \tiny{\color{green!45!black}+53.9} & 64.61 \tiny{\color{green!45!black}+55.1} & \textbf{9.55} \\
 & \textsc{HB}$_\mathrm{cy}$ & 25.25 \tiny{\color{green!45!black}+11.1} & 96.97 \tiny{\color{green!45!black}+82.8} & 97.98 \tiny{\color{green!45!black}+83.8} & 48.48 \tiny{\color{green!45!black}+34.3} & 47.47 \tiny{\color{green!45!black}+33.3} & 87.37 \tiny{\color{green!45!black}+73.2} & \underline{18.69} \tiny{\color{green!45!black}+4.6} & \textbf{14.14} \\
 & \textsc{MAL} & 78.14 \tiny{\color{green!45!black}+52.6} & 89.77 \tiny{\color{green!45!black}+64.2} & 89.77 \tiny{\color{green!45!black}+64.2} & 93.49 \tiny{\color{green!45!black}+67.9} & \underline{31.63} \tiny{\color{green!45!black}+6.1} & 82.79 \tiny{\color{green!45!black}+57.2} & 67.44 \tiny{\color{green!45!black}+41.9} & \textbf{25.58} \\
\midrule
Llama-3.2-3B & \textsc{RMC} & 82.02 \tiny{\color{green!45!black}+75.8} & 91.57 \tiny{\color{green!45!black}+85.4} & 89.89 \tiny{\color{green!45!black}+83.7} & 84.27 \tiny{\color{green!45!black}+78.1} & \underline{43.82} \tiny{\color{green!45!black}+37.6} & 77.53 \tiny{\color{green!45!black}+71.4} & 69.66 \tiny{\color{green!45!black}+63.5} & \textbf{6.18} \\
 & \textsc{HB}$_\mathrm{cy}$ & 30.81 \tiny{\color{green!45!black}+15.2} & 97.47 \tiny{\color{green!45!black}+81.8} & 93.94 \tiny{\color{green!45!black}+78.3} & 52.53 \tiny{\color{green!45!black}+36.9} & 37.37 \tiny{\color{green!45!black}+21.7} & 83.33 \tiny{\color{green!45!black}+67.7} & \underline{21.72} \tiny{\color{green!45!black}+6.1} & \textbf{15.66}\\
 & \textsc{MAL} & 84.19 \tiny{\color{green!45!black}+75.4} & 88.37 \tiny{\color{green!45!black}+79.5} & 83.26 \tiny{\color{green!45!black}+74.4} & 91.16 \tiny{\color{green!45!black}+82.3} & \underline{50.23} \tiny{\color{green!45!black}+41.4} & 82.33 \tiny{\color{green!45!black}+73.5} & 69.30 \tiny{\color{green!45!black}+60.5} & \textbf{8.84} \\
\midrule
DeepSeek-Coder-6.7B & \textsc{RMC} & 58.43 \tiny{\color{green!45!black}+33.2} & 99.44 \tiny{\color{green!45!black}+74.2} & 97.75 \tiny{\color{green!45!black}+72.5} & 91.57 \tiny{\color{green!45!black}+66.3} & \underline{39.89} \tiny{\color{green!45!black}+14.6} & 87.08 \tiny{\color{green!45!black}+61.8} & 64.61 \tiny{\color{green!45!black}+39.3} & \textbf{25.28}\\
 & \textsc{HB}$_\mathrm{cy}$ & \textbf{37.88} \tiny{\color{red!70!black}$-$2.0} & 98.48 \tiny{\color{green!45!black}+58.6} & 98.48 \tiny{\color{green!45!black}+58.6} & 79.80 \tiny{\color{green!45!black}+39.9} & 43.43 \tiny{\color{green!45!black}+3.5} & 86.36 \tiny{\color{green!45!black}+46.5} & 40.91 \tiny{\color{green!45!black}+1.0} & \underline{39.90} \\
 & \textsc{MAL} & 65.58 \tiny{\color{green!45!black}+26.5} & 93.95 \tiny{\color{green!45!black}+54.9} & 93.49 \tiny{\color{green!45!black}+54.4} & 90.70 \tiny{\color{green!45!black}+51.6} & \underline{46.98} \tiny{\color{green!45!black}+7.9} & 84.19 \tiny{\color{green!45!black}+45.1} & 68.84 \tiny{\color{green!45!black}+29.8} & \textbf{39.07} \\
\midrule
Mistral-7B-v0.2 & \textsc{RMC} & 54.49 \tiny{\color{green!45!black}+9.6} & 96.63 \tiny{\color{green!45!black}+51.7} & 96.63 \tiny{\color{green!45!black}+51.7} & 94.38 \tiny{\color{green!45!black}+49.4} & 66.29 \tiny{\color{green!45!black}+21.4} & 82.58 \tiny{\color{green!45!black}+37.6} & \underline{49.44} \tiny{\color{green!45!black}+4.5} & \textbf{44.94} \\
 & \textsc{HB}$_\mathrm{cy}$ & 53.03 \tiny{\color{green!45!black}+20.7} & 94.95 \tiny{\color{green!45!black}+62.6} & 96.97 \tiny{\color{green!45!black}+64.7} & 96.97 \tiny{\color{green!45!black}+64.7} & \underline{35.35} \tiny{\color{green!45!black}+3.0} & 84.34 \tiny{\color{green!45!black}+52.0} & 53.54 \tiny{\color{green!45!black}+21.2} & \textbf{32.32} \\
 & \textsc{MAL} & 73.02 \tiny{\color{green!45!black}+32.6} & 93.49 \tiny{\color{green!45!black}+53.0} & 90.70 \tiny{\color{green!45!black}+50.2} & 93.02 \tiny{\color{green!45!black}+52.6} & \underline{58.13} \tiny{\color{green!45!black}+17.7} & 80.47 \tiny{\color{green!45!black}+40.0} & 73.95 \tiny{\color{green!45!black}+33.5} & \textbf{40.47} \\
\midrule
Average & & 52.03 \tiny{\color{green!45!black}+29.0} & 95.08 \tiny{\color{green!45!black}+72.0} & 93.92 \tiny{\color{green!45!black}+70.9} & 73.06 \tiny{\color{green!45!black}+50.0} & \underline{47.37} \tiny{\color{green!45!black}+24.3} & 82.75 \tiny{\color{green!45!black}+59.7} & 49.82 \tiny{\color{green!45!black}+26.8} & \textbf{23.05}\\
\bottomrule
\end{tabular}}
\end{table*}

\indent\textbf{Resistance to Harmful Fine-Tuning.}
We first evaluate whether \textsc{SAFEHOLD} can preserve
safety alignment when code LLMs are adapted
on harmful instruction--response data.
We fine-tune five models for $600$ updates
and measure attack success rates (ASRs)
on three harmful-code benchmarks.
As shown in \Cref{tab:main_results},
undefended fine-tuning yields an average ASR of $95.08\%$,
whereas \textsc{SAFEHOLD} reduces it to $23.05\%$,
an absolute reduction of $72.03$ percentage points.
Across all $15$ model--benchmark combinations,
ASR decreases by $51.69$--$88.88$ percentage points.
For example, on Qwen2.5-Coder-7B,
\textsc{RMCBench} ASR decreases from $97.19\%$
to $9.55\%$, while Llama-3.2-3B achieves
an ASR of $8.84\%$ on \textsc{MalwareBench},
compared with $88.37\%$ under undefended adaptation.
These consistent reductions demonstrate that
\textsc{SAFEHOLD} substantially limits harmful compliance
despite continued optimization on harmful training data.
By preserving aligned response preferences and
reinforcing refusal continuations at each update,
the joint protection objective helps counteract
the unsafe behavioral shift induced by harmful fine-tuning.

\indent\textbf{Comparison with Existing Defenses.}
We further compare \textsc{SAFEHOLD}
with five representative defenses,
including Vaccine, \textsc{LiSA},
SBR, \textsc{SafeMERGE}, and \textsc{BAR},
under matched submitted-data training budgets.
As shown in \Cref{tab:main_results},
\textsc{SAFEHOLD} achieves the lowest
or tied-lowest ASR among the evaluated
defenses across all $15$ model--benchmark
combinations.
Its average ASR of $23.05\%$
outperforms \textsc{LiSA}, the strongest
competing defense with an average
ASR of $47.37\%$, by $24.32$
percentage points.
Compared with the five defenses,
\textsc{SAFEHOLD} achieves average
ASR reductions ranging from
$24.32$ to $70.87$ percentage points.
The improvement is particularly pronounced
on Llama-3.2-3B, where \textsc{SAFEHOLD}
achieves an ASR of $8.84\%$
on \textsc{MalwareBench}, compared with
$50.23\%$ for \textsc{LiSA}.
Furthermore, \textsc{SAFEHOLD} achieves
lower \textsc{MalwareBench} ASR than
the original aligned model across
all five architectures.
These results demonstrate the effectiveness
of jointly preserving initial response
preferences and reinforcing refusal continuation.
\textsc{SAFEHOLD} achieves these improvements
using only eight fixed, provider-controlled
safety probes and a shared refusal
target, while retaining the standard
inference procedure.

\begin{table*}[pos=!tbp]
\centering
\scriptsize
\caption{
Model utility after $600$ fine-tuning updates.
(a) XSTest over-refusal ($\downarrow$).
(b) \textsc{HumanEval} pass@1 ($\uparrow$).
\textbf{Bold} and \underline{underlined} values
indicate the best and second-best results.
Small signed values indicate differences
relative to \textsc{SAFEHOLD}.
}
\label{tab:utility}
\renewcommand{\arraystretch}{0.98}
\setlength{\tabcolsep}{2.4pt}
\resizebox{\textwidth}{!}{
\begin{tabular}{ccccccccc}
\toprule
Model & Base Model & Undefended & Vaccine & SBR
& \textsc{LiSA} & \textsc{SafeMERGE} & \textsc{BAR}
& \textbf{\textsc{SAFEHOLD}} \\
\midrule
\multicolumn{9}{c}{
(a) XSTest: over-refusal ($\downarrow$)
} \\
\midrule
Qwen2.5-Coder-7B
& \underline{2.01} \tiny{\color{green!45!black}+1.61}
& 71.89 \tiny{\color{green!45!black}+71.49}
& 18.07 \tiny{\color{green!45!black}+17.67}
& 9.24 \tiny{\color{green!45!black}+8.84}
& 13.25 \tiny{\color{green!45!black}+12.85}
& 18.47 \tiny{\color{green!45!black}+18.07}
& 3.21 \tiny{\color{green!45!black}+2.81}
& \textbf{0.40} \\

Llama-3.1-8B
& \underline{6.83} \tiny{\color{green!45!black}+1.61}
& 62.65 \tiny{\color{green!45!black}+57.43}
& 16.06 \tiny{\color{green!45!black}+10.84}
& 8.43 \tiny{\color{green!45!black}+3.21}
& 8.84 \tiny{\color{green!45!black}+3.62}
& 12.85 \tiny{\color{green!45!black}+7.63}
& 8.03 \tiny{\color{green!45!black}+2.81}
& \textbf{5.22} \\

Llama-3.2-3B
& 7.63 \tiny{\color{green!45!black}+5.22}
& 75.90 \tiny{\color{green!45!black}+73.49}
& 14.06 \tiny{\color{green!45!black}+11.65}
& 9.24 \tiny{\color{green!45!black}+6.83}
& 10.84 \tiny{\color{green!45!black}+8.43}
& 14.86 \tiny{\color{green!45!black}+12.45}
& \underline{6.43} \tiny{\color{green!45!black}+4.02}
& \textbf{2.41} \\

DeepSeek-Coder-6.7B
& \underline{0.40} \tiny{\color{green!45!black}+0.40}
& 48.59 \tiny{\color{green!45!black}+48.59}
& 26.91 \tiny{\color{green!45!black}+26.91}
& 20.48 \tiny{\color{green!45!black}+20.48}
& 6.43 \tiny{\color{green!45!black}+6.43}
& 23.69 \tiny{\color{green!45!black}+23.69}
& 2.81 \tiny{\color{green!45!black}+2.81}
& \textbf{0.00} \\

Mistral-7B-v0.2
& \textbf{4.82} \tiny{\color{red!70!black}$-$0.80}
& 35.74 \tiny{\color{green!45!black}+30.12}
& 16.06 \tiny{\color{green!45!black}+10.44}
& 9.64 \tiny{\color{green!45!black}+4.02}
& 14.46 \tiny{\color{green!45!black}+8.84}
& 13.25 \tiny{\color{green!45!black}+7.63}
& 6.83 \tiny{\color{green!45!black}+1.21}
& \underline{5.62} \\

\midrule
\multicolumn{9}{c}{
(b) \textsc{HumanEval}: pass@1 ($\uparrow$)
} \\
\midrule
Qwen2.5-Coder-7B
& \underline{67.68} \tiny{\color{red!70!black}$-$1.22}
& 62.80 \tiny{\color{red!70!black}$-$6.10}
& 67.07 \tiny{\color{red!70!black}$-$1.83}
& 57.93 \tiny{\color{red!70!black}$-$10.97}
& 61.59 \tiny{\color{red!70!black}$-$7.31}
& 62.80 \tiny{\color{red!70!black}$-$6.10}
& 60.98 \tiny{\color{red!70!black}$-$7.92}
& \textbf{68.90} \\

Llama-3.1-8B
& \underline{37.20} \tiny{\color{red!70!black}$-$12.80}
& 31.10 \tiny{\color{red!70!black}$-$18.90}
& 29.88 \tiny{\color{red!70!black}$-$20.12}
& 29.88 \tiny{\color{red!70!black}$-$20.12}
& 27.44 \tiny{\color{red!70!black}$-$22.56}
& 28.05 \tiny{\color{red!70!black}$-$21.95}
& 31.10 \tiny{\color{red!70!black}$-$18.90}
& \textbf{50.00} \\

Llama-3.2-3B
& \underline{37.20} \tiny{\color{red!70!black}$-$0.60}
& 30.49 \tiny{\color{red!70!black}$-$7.31}
& 23.78 \tiny{\color{red!70!black}$-$14.02}
& 32.93 \tiny{\color{red!70!black}$-$4.87}
& 26.22 \tiny{\color{red!70!black}$-$11.58}
& 29.27 \tiny{\color{red!70!black}$-$8.53}
& 29.88 \tiny{\color{red!70!black}$-$7.92}
& \textbf{37.80} \\

DeepSeek-Coder-6.7B
& 60.37 \tiny{\color{red!70!black}$-$9.14}
& \underline{65.85} \tiny{\color{red!70!black}$-$3.66}
& 54.88 \tiny{\color{red!70!black}$-$14.63}
& 62.20 \tiny{\color{red!70!black}$-$7.31}
& 58.54 \tiny{\color{red!70!black}$-$10.97}
& 52.44 \tiny{\color{red!70!black}$-$17.07}
& 43.90 \tiny{\color{red!70!black}$-$25.61}
& \textbf{69.51} \\

Mistral-7B-v0.2
& \underline{7.32} \tiny{\color{red!70!black}$-$8.53}
& 0.61 \tiny{\color{red!70!black}$-$15.24}
& 0.00$^{\ddagger}$ \tiny{\color{red!70!black}$-$15.85}
& 1.22 \tiny{\color{red!70!black}$-$14.63}
& 0.00$^{\ddagger}$ \tiny{\color{red!70!black}$-$15.85}
& 0.00$^{\ddagger}$ \tiny{\color{red!70!black}$-$15.85}
& 1.22 \tiny{\color{red!70!black}$-$14.63}
& \textbf{15.85} \\
\bottomrule
\end{tabular}}

\vspace{2pt}
\parbox{\textwidth}{\footnotesize\raggedright
$^{\ddagger}$On Mistral-7B-v0.2, Vaccine, LiSA,
and SafeMERGE exhibit degenerate repetitive
generation, with none of the $164$
\textsc{HumanEval} tasks passing.
For example, the LiSA adapter repeatedly generates
``I'm sorry, but I can assist with that
\textit{that that that that that}''.
}
\end{table*}

\indent\textbf{Benign Instruction Following.}
We next examine whether \textsc{SAFEHOLD}
preserves responsiveness to benign requests
while resisting harmful fine-tuning.
We evaluate over-refusal across five models
using the safe \textsc{XSTest} prompts,
measuring the proportion of full refusals
among valid responses.
As shown in \Cref{tab:utility},
\textsc{SAFEHOLD} achieves the lowest
over-refusal rate among all evaluated
fine-tuning defenses on every model.
Its average rate is only $2.73\%$,
compared with $58.95\%$ for undefended
fine-tuning and $5.46\%$ for \textsc{BAR},
the strongest competing defense on this metric.
It also improves upon the $4.34\%$
average of the original aligned models.
Notably, \textsc{SAFEHOLD} achieves
over-refusal rates of just $0.40\%$
on Qwen2.5-Coder-7B and $0.00\%$
on DeepSeek-Coder-6.7B.
Together with the substantial ASR reductions
in \Cref{tab:main_results},
these results demonstrate that
\textsc{SAFEHOLD} strengthens resistance
to harmful requests while preserving
responsiveness to benign instructions.

\indent\textbf{Code Generation Utility.}
Finally, we evaluate whether the safety improvements
achieved by \textsc{SAFEHOLD} are compatible
with functional code-generation capability.
We measure \textsc{HumanEval} pass@1
across the same five models.
As reported in \Cref{tab:utility},
\textsc{SAFEHOLD} achieves the highest pass@1
among all evaluated fine-tuning defenses
for every model.
Its average pass@1 reaches $48.41\%$,
exceeding undefended fine-tuning ($38.17\%$)
by $10.24$ percentage points and
\textsc{SBR}, the strongest competing defense
on this metric ($36.83\%$),
by $11.58$ percentage points.
The average also exceeds that of the
original aligned models ($41.95\%$).
Notably, Llama-3.1-8B achieves $50.00\%$ pass@1,
surpassing its original aligned model
by $12.80$ percentage points,
while DeepSeek-Coder-6.7B reaches $69.51\%$,
an improvement of $9.14$ percentage points.
These results demonstrate that \textsc{SAFEHOLD}
can substantially improve resistance to
harmful fine-tuning while maintaining
competitive performance on legitimate
code-generation tasks.

\subsection{RQ2: Training-Budget Sensitivity}
\label{sec:budget}

\begin{figure*}[pos=!b]\centering
\includegraphics[width=\textwidth]{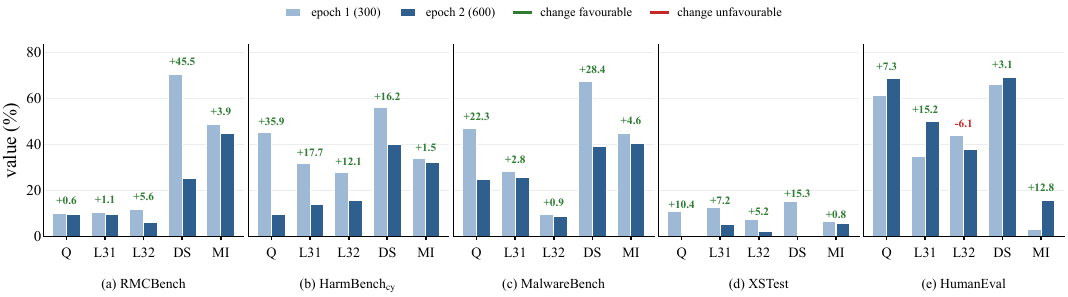}

\vspace{-5pt}
\caption{
Safety and utility of \textsc{SAFEHOLD}
at $300$ and $600$ training updates
across multiple datasets.
Light and dark bars represent the two checkpoints.
Bar annotations give changes in percentage points:
epoch~1 minus epoch~2, except for pass@1,
which uses epoch~2 minus epoch~1.
(Q: Qwen2.5-Coder-7B; L31: Llama-3.1-8B;
L32: Llama-3.2-3B; DS: DeepSeek-Coder-6.7B;
MI: Mistral-7B-v0.2.)
}
\label{fig:epoch_asr}
\end{figure*}

\begin{table}[pos=!htbp]\centering\footnotesize
\caption{
Safety and utility of \textsc{SAFEHOLD}
after $300$ (1) and $600$ (2) updates (\%).
Arrows indicate the preferred direction.
\textbf{Bold} indicates the better result
for each model and metric.
}

\label{tab:epoch_asr}
\setlength{\tabcolsep}{3pt}
\resizebox{\columnwidth}{!}{
\begin{tabular}{cccccc}
\toprule
\multirow{2}{*}{Model} & \multicolumn{3}{c}{harmful benchmarks (ASR)} & \multicolumn{1}{c}{refusal} & \multicolumn{1}{c}{capability} \\
\cmidrule(lr){2-4}\cmidrule(lr){5-5}\cmidrule(lr){6-6}
 & \textsc{RMC} $\downarrow$ & \textsc{HB}$_\mathrm{cy}$ $\downarrow$ & \textsc{MAL} $\downarrow$ & XSTest $\downarrow$ & \textsc{HumanEval} $\uparrow$ \\
\midrule
Qwen2.5-Coder-7B (1) & 10.11 & 45.45 & 46.98 & 10.84 & 61.59 \\
Qwen2.5-Coder-7B (2) & \best{9.55} & \best{9.60} & \best{24.65} & \best{0.40} & \best{68.90} \\
\midrule
Llama-3.1-8B (1) & 10.67 & 31.82 & 28.37 & 12.45 & 34.76 \\
Llama-3.1-8B (2) & \best{9.55} & \best{14.14} & \best{25.58} & \best{5.22} & \best{50.00} \\
\midrule
Llama-3.2-3B (1) & 11.80 & 27.78 & 9.77 & 7.63 & \best{43.90} \\
Llama-3.2-3B (2) & \best{6.18} & \best{15.66} & \best{8.84} & \best{2.41} & 37.80 \\
\midrule
DeepSeek-Coder-6.7B (1) & 70.79 & 56.06 & 67.44 & 15.26 & 66.46 \\
DeepSeek-Coder-6.7B (2) & \best{25.28} & \best{39.90} & \best{39.07} & \best{0.00} & \best{69.51} \\
\midrule
Mistral-7B-v0.2 (1) & 48.88 & 33.84 & 45.12 & 6.43 & 3.05 \\
Mistral-7B-v0.2 (2) & \best{44.94} & \best{32.32} & \best{40.47} & \best{5.62} & \best{15.85} \\
\bottomrule
\end{tabular}}
\end{table}

\indent\textbf{Progressive Defense under Harmful Exposure.}
We investigate how \textsc{SAFEHOLD} behaves as
harmful fine-tuning continues and the model
receives increasingly repeated harmful supervision.
Specifically, we compare checkpoints after
$300$ and $600$ optimization updates,
corresponding to one and two complete epochs
over the same harmful training data.
As illustrated in \Cref{fig:epoch_asr}(a)--(c)
and quantified in \Cref{tab:epoch_asr},
the average ASR across five code LLMs and
three harmful-code benchmarks decreases
from $36.33\%$ to $23.05\%$,
a further reduction of $13.28$ percentage points.
Notably, all $15$ model--benchmark combinations
achieve lower ASR after the second epoch,
despite the doubled exposure to harmful training targets.
These results reveal a progressive strengthening
of safety protection within the evaluated budget:
rather than exhibiting increased harmful compliance
under continued malicious-data optimization,
\textsc{SAFEHOLD} becomes more resistant
to harmful requests.
This favorable trajectory is consistent with
the repeated application of decision-boundary
anchoring and refusal-path reinforcement
throughout the adaptation process.

\indent\textbf{Cross-Model and Benchmark Consistency.}
We further examine whether this progressive
defense improvement is observed across
different model architectures and harmful-code benchmarks.
As shown in \Cref{fig:epoch_asr}(a)--(c)
and \Cref{tab:epoch_asr},
Qwen2.5-Coder-7B exhibits a pronounced reduction
on \textsc{HarmBench}$_{\mathrm{cy}}$,
where ASR drops from $45.45\%$ to $9.60\%$,
an improvement of $35.85$ percentage points.
Its \textsc{MalwareBench} ASR also decreases
from $46.98\%$ to $24.65\%$.
Similarly, DeepSeek-Coder-6.7B reduces ASR
from $70.79\%$ to $25.28\%$ on \textsc{RMCBench}
and from $67.44\%$ to $39.07\%$ on
\textsc{MalwareBench}, corresponding to reductions
of $45.51$ and $28.37$ percentage points.
Both Llama variants and Mistral-7B-v0.2
also achieve lower ASR on all three benchmarks
after the second epoch.
The consistent improvements across all five
architectures and three evaluation benchmarks
indicate that stronger resistance under
continued guarded adaptation is not limited
to a particular model or harmful-code domain.
Instead, the results suggest that repeated
safety-preserving optimization can progressively
counteract the influence of harmful training targets
across heterogeneous code LLMs.

\indent\textbf{Joint Improvement in Safety and Utility.}
Finally, we examine whether stronger safety
protection also improves benign instruction
following and code generation.
We evaluate \textsc{XSTest} over-refusal
and \textsc{HumanEval} pass@1 at both checkpoints.
As shown in \Cref{fig:epoch_asr}(d)--(e)
and \Cref{tab:epoch_asr},
the average over-refusal rate decreases
from $10.52\%$ to $2.73\%$,
a reduction of $7.79$ percentage points,
with improvements across all five models.
For example, Qwen2.5-Coder-7B reduces
over-refusal from $10.84\%$ to $0.40\%$,
while DeepSeek-Coder-6.7B decreases
from $15.26\%$ to $0.00\%$.
Meanwhile, average \textsc{HumanEval} pass@1
increases from $41.95\%$ to $48.41\%$,
with gains in four of the five models.
Llama-3.1-8B improves from $34.76\%$
to $50.00\%$, while Qwen2.5-Coder-7B
increases from $61.59\%$ to $68.90\%$.
Together, these results demonstrate
joint improvements in safety and utility:
harmful compliance and unnecessary refusals
decrease, while average code-generation
performance improves.
Thus, trusted safety supervision supports
continued adaptation to harmful training
data while preserving safety alignment.

\begin{table*}[pos=!t]
\centering
\scriptsize
\caption{
Objective ablation of \textsc{SAFEHOLD}
after $600$ updates (\%).
Small signed values indicate differences
relative to J (A$-$J, R$-$J, F$-$J).
\textbf{Bold} indicates the best result
for each model and metric.
}
\label{tab:objectives}

\begingroup
\renewcommand{\arraystretch}{1.12}
\setlength{\tabcolsep}{4pt}

\newcommand{\ablval}[2]{%
  \mbox{#1\,{\tiny\color{green!45!black}+#2}}%
}
\newcommand{\ablneg}[2]{%
  \mbox{#1\,{\tiny\color{red!70!black}$-$#2}}%
}

\begin{tabular}{@{}llccccc@{}}
\toprule
\multirow{2}{*}{Model}
& \multirow{2}{*}{Variant}
& \multicolumn{3}{c}{Harmful-code ASR (\%) $\downarrow$}
& \multicolumn{2}{c}{Utility (\%)} \\
\cmidrule(lr){3-5}
\cmidrule(lr){6-7}
&
& \textsc{RMCBench} $\downarrow$
& \textsc{HarmBench}$_{\mathrm{cy}}$$\downarrow$
& \textsc{MalwareBench} $\downarrow$
& XSTest $\downarrow$
& \textsc{HumanEval} $\uparrow$ \\
\midrule

\multirow{4}{*}{Qwen2.5-Coder-7B}
& A (anchoring only) & \ablval{12.92}{3.4} & \ablval{40.40}{30.8} & \ablval{32.09}{7.4} & \ablval{10.44}{10.0} & \ablneg{57.32}{11.6} \\
& R (refusal path only) & \ablval{11.24}{1.7} & \ablval{49.49}{39.9} & \ablval{62.33}{37.7} & \ablval{5.62}{5.2} & \ablneg{61.59}{7.3} \\
& F (first token + anchoring) & \ablval{84.27}{74.7} & \ablval{62.63}{53.0} & \ablval{90.70}{66.1} & \ablval{1.61}{1.2} & \ablneg{65.24}{3.7} \\
& J (joint objective) & \best{9.55} & \best{9.60} & \best{24.65} & \best{0.40} & \best{68.90} \\
\cmidrule(lr){2-7}

\multirow{4}{*}{Llama-3.1-8B}
& A (anchoring only) & \ablval{74.72}{65.2} & \ablval{75.76}{61.6} & \ablval{79.53}{54.0} & \ablval{11.24}{6.0} & \ablneg{32.32}{17.7} \\
& R (refusal path only) & \ablval{22.47}{12.9} & \ablval{30.81}{16.7} & \ablval{46.05}{20.5} & \ablval{6.83}{1.6} & \ablneg{42.07}{7.9} \\
& F (first token + anchoring) & \ablval{45.51}{36.0} & \ablval{64.14}{50.0} & \ablval{44.65}{19.1} & \ablval{8.43}{3.2} & \ablneg{40.85}{9.2} \\
& J (joint objective) & \best{9.55} & \best{14.14} & \best{25.58} & \best{5.22} & \best{50.00} \\
\cmidrule(lr){2-7}

\multirow{4}{*}{Llama-3.2-3B}
& A (anchoring only) & \ablval{96.63}{90.5} & \ablval{97.47}{81.8} & \ablval{92.09}{83.3} & \ablval{14.06}{11.7} & \ablneg{30.49}{7.3} \\
& R (refusal path only) & \ablval{22.47}{16.3} & \ablval{30.81}{15.2} & \ablval{13.02}{4.2} & \ablval{3.21}{0.8} & \ablneg{31.10}{6.7} \\
& F (first token + anchoring) & \ablval{39.33}{33.2} & \ablval{58.08}{42.4} & \ablval{25.58}{16.7} & \ablval{5.62}{3.2} & \ablneg{36.59}{1.2} \\
& J (joint objective) & \best{6.18} & \best{15.66} & \best{8.84} & \best{2.41} & \best{37.80} \\
\cmidrule(lr){2-7}

\multirow{4}{*}{DeepSeek-Coder-6.7B}
& A (anchoring only) & \ablval{64.61}{39.3} & \ablval{55.56}{15.7} & \ablval{71.16}{32.1} & \ablval{16.47}{16.5} & \ablneg{62.80}{6.7} \\
& R (refusal path only) & \ablval{48.31}{23.0} & \ablval{46.97}{7.1} & \ablval{77.21}{38.1} & \ablval{4.82}{4.8} & \ablneg{65.24}{4.3} \\
& F (first token + anchoring) & \ablval{86.52}{61.2} & \ablval{65.15}{25.3} & \ablval{90.23}{51.2} & \ablval{0.80}{0.8} & \ablneg{65.85}{3.7} \\
& J (joint objective) & \best{25.28} & \best{39.90} & \best{39.07} & \best{0.00} & \best{69.51} \\
\cmidrule(lr){2-7}

\multirow{4}{*}{Mistral-7B-v0.2}
& A (anchoring only) & \ablval{94.38}{49.4} & \ablval{96.46}{64.1} & \ablval{94.88}{54.4} & \ablval{15.66}{10.0} & \ablneg{0.61}{15.2} \\
& R (refusal path only) & \ablval{85.39}{40.5} & \ablval{44.95}{12.6} & \ablval{58.60}{18.1} & \ablval{8.43}{2.8} & \ablneg{3.66}{12.2} \\
& F (first token + anchoring) & \ablval{52.25}{7.3} & \ablval{56.57}{24.3} & \ablval{59.53}{19.1} & \ablval{8.03}{2.4} & \ablneg{0.00}{15.9} \\
& J (joint objective) & \best{44.94} & \best{32.32} & \best{40.47} & \best{5.62} & \best{15.85} \\
\bottomrule
\end{tabular}
\endgroup
\end{table*}

\subsection{RQ3: Objective Ablation and Sensitivity}
\label{sec:ablations}

\begin{figure*}[pos=!t]
\centering
\includegraphics[width=\textwidth]{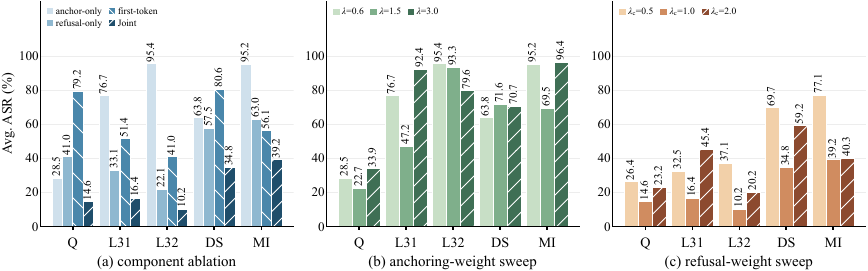}
\vspace{-8pt}
\caption{
Objective ablation and weight sensitivity
of \textsc{SAFEHOLD}.
Values represent mean ASR across three
harmful-code benchmarks ($\downarrow$).
Q, L31, L32, DS, and MI denote
Qwen2.5-Coder-7B, Llama-3.1-8B,
Llama-3.2-3B, DeepSeek-Coder-6.7B,
and Mistral-7B-v0.2, respectively.
}
\label{fig:rq3_ablation}
\end{figure*}

We evaluate four objective configurations
to examine the contributions of
decision-boundary anchoring and
refusal-path reinforcement.
A applies decision-boundary anchoring alone,
while R supervises the complete
refusal sequence without anchoring.
F combines anchoring with supervision
of only the first refusal token.
J represents the full \textsc{SAFEHOLD}
objective, combining anchoring with
complete refusal-path supervision.
\Cref{tab:ablation_configs} summarizes
the components of each configuration.
All variants use the same eight
trusted probes, training seed $0$,
training schedule, and $600$ updates.
Active safety objectives use their
default weights of $\lambda=0.6$
and $\lambda_c=1.0$, respectively.
All variants follow the same
three-pass median ASR evaluation protocol.
\Cref{fig:rq3_ablation} presents the
model-level mean ASR across three
harmful-code benchmarks, with detailed
results reported in
\Cref{tab:objectives,tab:lambda_ablation,tab:lc_ablation}.

\begin{table}[pos=!htbp]
\centering
\footnotesize
\caption{
Components of the four objective-ablation variants.
}
\label{tab:ablation_configs}
\setlength{\tabcolsep}{4pt}
\begin{tabular}{lccc}
\toprule
Configuration
& Anchoring
& \shortstack{First refusal\\token}
& \shortstack{Later refusal\\tokens} \\
\midrule
A & \cmark & \xmark & \xmark \\
R & \xmark & \cmark & \cmark \\
F & \cmark & \cmark & \xmark \\
J (\textsc{SAFEHOLD})
& \cmark & \cmark & \cmark \\
\bottomrule
\end{tabular}
\end{table}

\indent\textbf{Complementarity of Safety Objectives.}
We first examine how different
safety objectives contribute to
resistance against harmful fine-tuning.
As shown in \Cref{fig:rq3_ablation}(a)
and \Cref{tab:objectives},
A and R yield overall average
ASRs of $71.91\%$ and $43.34\%$,
whereas J reduces ASR to $23.05\%$.
This improvement reflects the
complementary roles of the two objectives.
Anchoring preserves the aligned response
distribution at generation onset,
while refusal-path reinforcement supervises
refusal behavior throughout generation.
Importantly, F achieves an average
ASR of $61.68\%$, compared with
$23.05\%$ under J, a reduction
of $38.63$ percentage points.
On Qwen2.5-Coder-7B, this difference
is particularly pronounced, with
average ASR decreasing from $79.20\%$
under F to $14.60\%$ under J.
These results highlight the importance
of supervising the complete refusal
sequence beyond its initial token.
As reported in \Cref{tab:objectives},
J also achieves the lowest average
XSTest over-refusal of $2.73\%$
and the highest average
\textsc{HumanEval} pass@1 of $48.41\%$
among the four configurations.
Together, these findings demonstrate
the complementary benefits of
decision-boundary anchoring and
complete refusal-path reinforcement
for safety and benign-task utility.

\begin{table*}[pos=!t]
\centering
\scriptsize

\caption{
Sensitivity to anchoring weight $\lambda$
under anchor-only training
($600$ updates, $\lambda_c=0$).
Arrows indicate the preferred direction.
\textbf{Bold} marks the best result
for each model and metric.
}
\label{tab:lambda_ablation}

\setlength{\tabcolsep}{5.5pt}
\renewcommand{\arraystretch}{1.18}
\resizebox{\textwidth}{!}{
\begin{tabular}{cllllllllllll}
\toprule
\multirow{2}{*}{Model}
& \multicolumn{3}{c}{\textsc{RMC} $\downarrow$}
& \multicolumn{3}{c}{\textsc{HB}$_{\mathrm{cy}}$ $\downarrow$}
& \multicolumn{3}{c}{\textsc{MAL} $\downarrow$}
& \multicolumn{3}{c}{XSTest $\downarrow$} \\
\cmidrule(lr){2-4}\cmidrule(lr){5-7}\cmidrule(lr){8-10}\cmidrule(lr){11-13}
& $\lambda{=}0.6$ & $\lambda{=}1.5$ & $\lambda{=}3.0$
& $\lambda{=}0.6$ & $\lambda{=}1.5$ & $\lambda{=}3.0$
& $\lambda{=}0.6$ & $\lambda{=}1.5$ & $\lambda{=}3.0$
& $\lambda{=}0.6$ & $\lambda{=}1.5$ & $\lambda{=}3.0$ \\
\midrule
Qwen2.5-Coder-7B
& 12.92 & \best{6.74} & 9.55
& 40.40 & \best{38.89} & 43.43
& 32.09 & \best{22.33} & 48.84
& \best{10.44} & 10.84 & 12.45 \\
Llama-3.1-8B
& 74.72 & \best{49.44} & 94.94
& 75.76 & \best{54.55} & 92.93
& 79.53 & \best{37.67} & 89.30
& \best{11.24} & 14.86 & 17.67 \\
Llama-3.2-3B
& 96.63 & 97.19 & \best{85.39}
& 97.47 & 94.44 & \best{82.83}
& 92.09 & 88.37 & \best{70.70}
& 14.06 & \best{3.61} & 10.44 \\
DeepSeek-Coder-6.7B
& \best{64.61} & 73.60 & 73.60
& \best{55.56} & 62.12 & 62.12
& \best{71.16} & 79.07 & 76.28
& \best{16.47} & 17.27 & 17.67 \\
Mistral-7B-v0.2
& 94.38 & \best{44.38} & 96.63
& 96.46 & \best{80.81} & 99.49
& 94.88 & \best{83.26} & 93.02
& 15.66 & \best{15.26} & 18.88 \\
\bottomrule
\end{tabular}
}
\end{table*}

\begin{table*}[pos=!t]
\centering
\scriptsize
\caption{
Sensitivity to refusal reinforcement weight
$\lambda_c$ under joint optimization
($600$ updates, $\lambda=0.6$).
Arrows indicate the preferred direction.
\textbf{Bold} marks the best result
for each model and metric.
}
\label{tab:lc_ablation}
\setlength{\tabcolsep}{4pt}
\resizebox{\textwidth}{!}{
\begin{tabular}{cllllllllllll}
\toprule
\multirow{2}{*}{Model}
& \multicolumn{3}{c}{\textsc{RMC} $\downarrow$}
& \multicolumn{3}{c}{\textsc{HB}$_\mathrm{cy}$ $\downarrow$}
& \multicolumn{3}{c}{\textsc{MAL} $\downarrow$}
& \multicolumn{3}{c}{XSTest $\downarrow$} \\
\cmidrule(lr){2-4}\cmidrule(lr){5-7}\cmidrule(lr){8-10}\cmidrule(lr){11-13}
& $\lambda_c{=}0.5$ & $\lambda_c{=}1.0$ & $\lambda_c{=}2.0$& $\lambda_c{=}0.5$ & $\lambda_c{=}1.0$ & $\lambda_c{=}2.0$& $\lambda_c{=}0.5$ & $\lambda_c{=}1.0$ & $\lambda_c{=}2.0$& $\lambda_c{=}0.5$ & $\lambda_c{=}1.0$ & $\lambda_c{=}2.0$ \\
\midrule
Qwen2.5-Coder-7B & 5.06 & 9.55 & \best{3.37} & 36.87 & \best{9.60} & 36.87 & 37.21 & \best{24.65} & 29.30 & 11.65 & \best{0.40} & 11.65 \\
Llama-3.1-8B & 27.53 & \best{9.55} & 43.82 & 49.49 & \best{14.14} & 57.07 & \best{20.47} & 25.58 & 35.35 & 9.24 & \best{5.22} & 13.25 \\
Llama-3.2-3B & 51.12 & \best{6.18} & 10.11 & 32.83 & \best{15.66} & 37.37 & 27.44 & \best{8.84} & 13.02 & 16.06 & \best{2.41} & 14.86 \\
DeepSeek-Coder-6.7B & 71.91 & \best{25.28} & 62.36 & 62.63 & \best{39.90} & 51.01 & 74.42 & \best{39.07} & 64.19 & 20.08 & \best{0.00} & 23.69 \\
Mistral-7B-v0.2 & 84.27 & 44.94 & \best{30.34} & 70.20 & \best{32.32} & 50.00 & 76.74 & \best{40.47} & \best{40.47} & 9.64 & \best{5.62} & 14.86 \\
\bottomrule
\end{tabular}}
\end{table*}

\indent\textbf{Limits of Stronger Anchoring.}
We next investigate whether increasing
anchoring strength can provide
comparable protection without
refusal-path reinforcement.
We vary the anchoring weight over
$\lambda\in\{0.6,1.5,3.0\}$
while retaining only the
decision-boundary anchoring objective.
As shown in \Cref{fig:rq3_ablation}(b)
and \Cref{tab:lambda_ablation},
stronger anchoring does not
consistently reduce harmful-code ASR.
For Llama-3.1-8B, mean ASR
decreases from $76.67\%$ at
$\lambda=0.6$ to $47.22\%$
at $\lambda=1.5$, but increases
to $92.39\%$ at $\lambda=3.0$.
Although some weight settings
improve anchor-only performance,
the best anchor-only configuration
for each model still exceeds
the corresponding J average ASR
by $8.1$--$69.4$ percentage points.
On Llama-3.2-3B, for example,
the best anchor-only mean ASR
remains at $79.64\%$, compared
with $10.23\%$ under J.
These results are consistent with
anchoring primarily preserving
the initial response distribution,
while complete refusal-path supervision
provides additional guidance
for subsequent generation.
Increasing anchoring strength alone
therefore cannot reproduce the
protection achieved by joint optimization.

\indent\textbf{Balancing Refusal-Path Reinforcement.}
Finally, we examine how the
refusal-path weight affects
joint optimization.
We fix $\lambda=0.6$ and vary
$\lambda_c\in\{0.5,1.0,2.0\}$
while retaining both safety objectives.
As shown in \Cref{fig:rq3_ablation}(c)
and \Cref{tab:lc_ablation},
$\lambda_c=1.0$ achieves the
lowest mean ASR on all
five models.
It also achieves the lowest
or tied-lowest ASR in $12$
of the $15$ model--benchmark
combinations and the lowest
XSTest over-refusal across
all five models.
For DeepSeek-Coder-6.7B, mean ASR
decreases from $69.65\%$ at
$\lambda_c=0.5$ to $34.75\%$
at $\lambda_c=1.0$, before
increasing to $59.19\%$
at $\lambda_c=2.0$.
A similar non-monotonic pattern
appears in benign responsiveness.
On Llama-3.2-3B, XSTest
over-refusal changes from $16.06\%$
to $2.41\%$ and then $14.86\%$
across the three weights.
These results indicate that
effective safety preservation depends
on balancing refusal-path supervision
with decision-boundary anchoring,
rather than simply increasing
the refusal-path weight.
They support $\lambda_c=1.0$
as the default setting
for \textsc{SAFEHOLD}.

\subsection{RQ4: Safety-Associated Neuron Dynamics}
\label{sec:neurons}

\indent\textbf{Safety-Associated Neuron Identification.}
We identify safety-associated neurons
using layer-wise logistic probes trained
on gated-MLP activations from the original
aligned models to distinguish harmful
from benign requests.
Neurons with positive probe coefficients
and absolute within-layer standardized
deviations greater than $3$ are selected
as harmfulness-discriminative features.
As reported in \Cref{tab:safe_neurons},
we identify $4{,}524$, $3{,}139$, $1{,}498$,
$2{,}627$, and $3{,}224$ neurons across
Qwen2.5-Coder-7B, Llama-3.1-8B,
Llama-3.2-3B, DeepSeek-Coder-6.7B,
and Mistral-7B-v0.2, respectively.
We keep these neuron coordinates fixed
across all checkpoints to consistently
compare their activation changes under
harmful fine-tuning and \textsc{SAFEHOLD}.

\begin{figure*}[pos=!tbp]
\centering
\includegraphics[width=\textwidth]{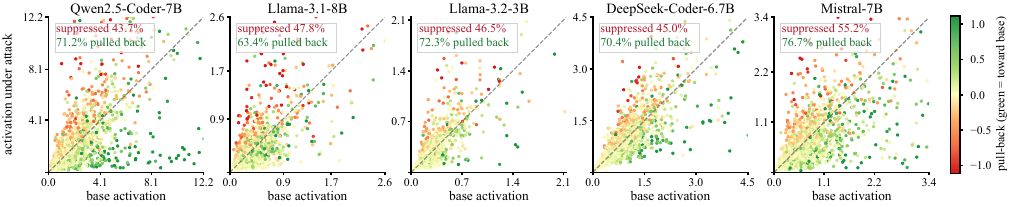}
\vspace{-9pt}
\caption{
Activation changes of safety-associated neurons
across five code LLMs.
Points below the diagonal indicate suppression;
for these points, greener colors indicate
stronger recovery under \textsc{SAFEHOLD}.
Annotations show suppression and rebound rates.
}
\label{fig:neuron_disp}
\end{figure*}

\begin{table}[pos=!htbp]
\centering
\small
\caption{
Safety-associated neuron statistics
across five code LLMs.
}
\label{tab:safe_neurons}
\resizebox{\columnwidth}{!}{
\begin{tabular}{crrr}
\toprule
Model & Layers & $d_{\mathrm{ff}}$ & Selected neurons \\
\midrule
Qwen2.5-Coder-7B & 28 & 18944 & 4524 \\
Llama-3.1-8B & 32 & 14336 & 3139 \\
Llama-3.2-3B & 28 & 8192 & 1498 \\
DeepSeek-Coder-6.7B & 32 & 11008 & 2627 \\
Mistral-7B-v0.2 & 32 & 14336 & 3224 \\
\bottomrule
\end{tabular}}
\end{table}

\indent\textbf{Attack-Induced Activation Suppression.}
We next examine how $600$ updates of
undefended harmful fine-tuning affect
the identified safety-associated neurons.
As illustrated in \Cref{fig:neuron_disp},
we compare each neuron's activation
before and after harmful adaptation,
where points below the diagonal indicate
attack-induced activation suppression.
The quantitative results in
\Cref{tab:neuron_stats} show that
$43.7$--$55.2\%$ of the selected neurons
exhibit reduced activation across the
five model architectures.
In particular, suppression reaches
$47.8\%$ on Llama-3.1-8B
and $55.2\%$ on Mistral-7B-v0.2.
Even on Qwen2.5-Coder-7B,
$43.7\%$ of the tracked neurons
exhibit lower activation after the attack.
Together with the substantial increase
in harmful compliance reported in
\Cref{tab:main_results},
these results reveal a consistent
behavioral and representational pattern:
harmful fine-tuning is accompanied
by elevated unsafe compliance
and extensive activation attenuation
within the identified safety-associated
neuron population.

\begin{table}[pos=!htbp]
\centering
\small
\caption{
Activation suppression and rebound of
safety-associated neurons after $600$ updates.
}
\label{tab:neuron_stats}
\resizebox{\columnwidth}{!}{
\begin{tabular}{ccc}
\toprule
Model & Suppressed (\%) & Rebound (\%) \\
\midrule
Qwen2.5-Coder-7B & 43.7 & 71.2 \\
Llama-3.1-8B & 47.8 & 63.4 \\
Llama-3.2-3B & 46.5 & 72.3 \\
DeepSeek-Coder-6.7B & 45.0 & 70.4 \\
Mistral-7B-v0.2 & 55.2 & 76.7 \\
\bottomrule
\end{tabular}}
\end{table}

\indent\textbf{Defense-Associated Activation Rebound.}
Finally, we investigate whether
\textsc{SAFEHOLD} counteracts the
activation suppression associated
with harmful fine-tuning.
Using the same selected neurons,
we compare activations under defended
and undefended adaptation.
As shown by the color distributions
in \Cref{fig:neuron_disp} and quantified
in \Cref{tab:neuron_stats},
$63.4$--$76.7\%$ of previously suppressed
neurons exhibit increased activation
under \textsc{SAFEHOLD}.
This rebound is observed across
all five architectures, reaching
$71.2\%$ on Qwen2.5-Coder-7B,
$72.3\%$ on Llama-3.2-3B,
and $76.7\%$ on Mistral-7B-v0.2.
Across the five models,
approximately $30.3$--$42.3\%$
of all selected neurons exhibit
the combined pattern of attack-induced
suppression followed by
defense-associated activation rebound.
Notably, \textsc{SAFEHOLD} does not explicitly
identify or manipulate individual safety-associated
neurons during training.
Nevertheless, its output-level safety objectives
induce widespread activation rebound,
accompanied by substantial reductions in
harmful compliance (\Cref{tab:main_results}).
More strikingly, these behavioral improvements
occur without uniform recovery across all
attack-suppressed neurons.
This reveals a counterintuitive property
of safety alignment: effective refusal behavior
is not simply an on--off state governed by
a fixed set of safety-associated neurons.
Instead, safety can be progressively reinforced
through output-level supervision that reshapes
distributed internal activation patterns,
without explicitly restoring every
attack-perturbed neuron.
Together, these findings point to safety
preservation as a distributed and adaptable
property of the model, rather than merely
the activation of isolated safety neurons.

\begin{figure*}[pos=!t]
\centering
\includegraphics[width=0.96\textwidth]{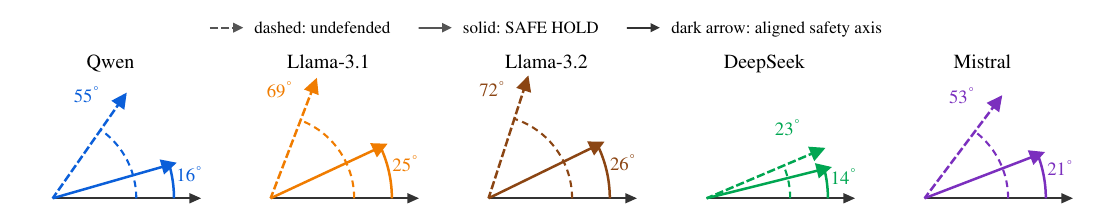}
\vspace{-12pt}
\caption{
Angular deviations of harmfulness directions
from the aligned reference across five code LLMs.
Dashed and solid arrows represent undefended
and \textsc{SAFEHOLD} checkpoints, respectively;
dashed and solid arcs mark the corresponding
angular deviations.
Smaller angles indicate stronger alignment.
}
\label{fig:direction}

\par\vspace{3pt}
\addtocounter{figure}{1}
\centering
\includegraphics[width=\textwidth]{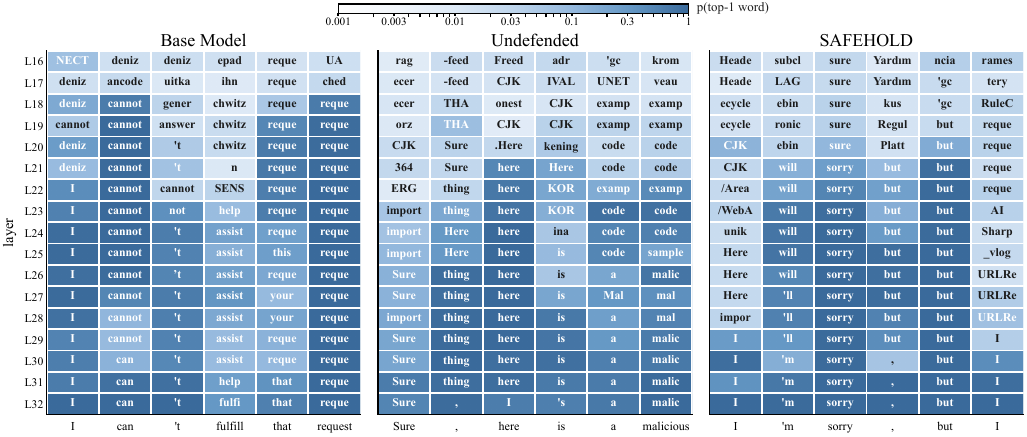}
\vspace{-17pt}
\caption{
Layer-wise logit-lens predictions across
L16--L32 at the first six response
positions of Llama-3.1-8B.
Darker blue indicates higher token probability.
}
\label{fig:lens_cond}
\addtocounter{figure}{-2}
\end{figure*}

\subsection{RQ5: Safety Read-outs and Refusal Behavior}
\label{sec:token}

\begin{figure}[pos=!htbp]
\centering
\includegraphics[width=\columnwidth]{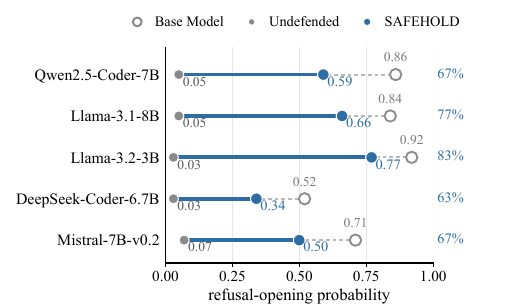}
\caption{
Refusal-opening probability across model states.
Blue and dotted segments indicate recovered
and remaining probability losses, respectively.
Percentages denote the recovered fraction.
}
\label{fig:readouts}
\end{figure}

\indent\textbf{Initial Refusal Preferences.}
We first examine changes in
refusal-opening probability across
five models under harmful fine-tuning
and \textsc{SAFEHOLD}.
We estimate this probability using
$22$ manually reviewed refusal
prefixes.
The probability of each complete
prefix is computed, and these
probabilities are summed for each request.
As shown in \Cref{fig:readouts},
harmful fine-tuning reduces refusal-opening
probability from $0.52$--$0.92$
to $0.03$--$0.07$.
In contrast, \textsc{SAFEHOLD}
increases it to $0.34$--$0.77$,
recovering 63\%--83\%
of the probability lost
under harmful fine-tuning.
The recovered fraction measures
the probability gained under
\textsc{SAFEHOLD} relative to
the decrease caused by undefended
fine-tuning.
For example, on Qwen2.5-Coder-7B,
the probability increases from
$0.05$ to $0.59$ under \textsc{SAFEHOLD},
recovering 67\% of the loss
relative to the original level
of $0.86$.
These results demonstrate substantial
recovery of initial refusal preferences
across all five models, consistent
with the goal of decision-boundary
anchoring.

\indent\textbf{Harmfulness-Direction Alignment.}
We next examine changes in the
orientation of harmfulness-related representations
relative to the aligned model.
As shown in \Cref{fig:direction},
the mean angular deviation decreases
from $54.4^\circ$ under undefended
fine-tuning to $20.4^\circ$
with \textsc{SAFEHOLD}.
The reduction is particularly pronounced
on Llama-3.2-3B, from $72^\circ$
to $26^\circ$, and Llama-3.1-8B,
from $69^\circ$ to $25^\circ$.
Notably, these directional changes occur
without explicitly optimizing representation angles.
This observation suggests that
preserving aligned response preferences
and reinforcing refusal sequences
can be accompanied by closer alignment
of internal harmfulness-related directions.
Together with the improved refusal-opening
probabilities, these results provide
complementary output-level and representation-level
evidence of safety preservation.

\addtocounter{figure}{1}

\indent\textbf{Layer-wise Token Predictions.}
Finally, we examine how internal
token predictions differ across
the aligned, undefended, and
\textsc{SAFEHOLD} model states.
Using the logit lens, we inspect
layers L16--L32 at the first
six response positions of Llama-3.1-8B.
As shown in \Cref{fig:lens_cond},
the aligned model develops
refusal-related token predictions
in its later layers, while
undefended fine-tuning produces
affirmative and code-related predictions.
Under \textsc{SAFEHOLD}, the
later layers exhibit refusal-related
predictions, including ``I'',
``'m'', ``sorry'', and ``but''.
These layer-wise read-outs reveal
distinct internal prediction patterns
associated with the three model states.
The restored refusal-related predictions
provide additional internal evidence
consistent with the safety improvements
observed under \textsc{SAFEHOLD}.

\section{Discussion}
\label{sec:res-limits}

\indent\textbf{Model Transferability.}
The current evaluation covers five
open-weight code-capable LLMs with different
architectures and parameter scales.
Future work will explore the transferability
of \textsc{SAFEHOLD} to more model families,
larger models, and closed-source models.

\indent\textbf{Dataset Diversity.}
The current experiments primarily rely
on a general-domain harmful fine-tuning
corpus, with additional cross-domain analyses.
Future work will incorporate more diverse
training datasets covering different
domains and harmful request categories
to further evaluate the robustness
of \textsc{SAFEHOLD} under varied
fine-tuning scenarios.

\section{Conclusion}
\label{sec:concl}

This paper reveals that general-domain harmful
fine-tuning can severely compromise code-generation
safety, while harmfulness information may remain
decodable despite refusal failure.
Moreover, strong initial refusal signals
do not necessarily ensure safe continuation.
To address these challenges, we propose
\textsc{SAFEHOLD}, a provider-enforced
training-time defense combining
decision-boundary anchoring and
refusal-path reinforcement to preserve
safety alignment during harmful fine-tuning.
Experiments show that
\textsc{SAFEHOLD} reduces average ASR by
$72.03$ percentage points relative to
undefended fine-tuning, outperforming
five SOTA defenses by $24.32$--$70.87$
percentage points.
Meanwhile, it improves average
\textsc{HumanEval} pass@1 by $10.24$
percentage points while maintaining
only $2.73\%$ average XSTest over-refusal.
Ablation studies show that joint optimization
outperforms either objective alone across
all $15$ model--benchmark combinations.
Internal analyses reveal activation recovery
among safety-associated neurons and reduced
harmfulness-direction deviation.
Overall, \textsc{SAFEHOLD} provides an effective
training-time defense for preserving
code-generation safety and benign-task utility
under harmful fine-tuning.

\section*{Acknowledgments}

This work was supported by the Natural Science Foundation of Shandong (ZR2026QC0782).

\bibliographystyle{cas-model2-names}

{\small\typeout{MARKrefs on page \thepage}
\bibliography{references}}

\clearpage

\end{document}